\documentclass[journal]{IEEEtran}
\usepackage{amsmath,amsfonts}
\usepackage{algorithmic}
\usepackage{algorithm}
\usepackage{array}
\usepackage[caption=false,font=normalsize,labelfont=sf,textfont=sf]{subfig}
\usepackage{textcomp}
\usepackage{stfloats}
\usepackage{pifont}
\usepackage{url}
\usepackage{verbatim}
\usepackage{graphicx}
\usepackage{cite}
\usepackage{pdfpages}
\usepackage{adjustbox}
\usepackage{caption}
\usepackage{multirow} 
\usepackage{booktabs} 
\usepackage{array}    
\usepackage{siunitx}  
\usepackage{xcolor}

\usepackage{hyperref}
\hypersetup{hypertex=true,
	colorlinks=true,
	linkcolor=blue,
	anchorcolor=blue,
	citecolor=blue}
\usepackage{threeparttable}

\begin{document}

\title{XMatchAD: A Cross-Modal Matching Perspective on Reconstruction-based Anomaly Detection}

\author{Mingxiu Cai, Zhe Zhang, Gaochang Wu,~\IEEEmembership{Member,~IEEE}, Tianyou Chai\hspace{-1.0mm},~\IEEEmembership{Life Fellow,~IEEE}
\thanks{This work is supported in part by the Science and Technology Major Project of Liaoning Province under Grant No. 2024JH1/11700048, in part by the Research Program of the Liaoning Liaohe Laboratory under Grant No. LLL23ZZ-05-01, in part by the Natural Science Foundation of Liaoning Province under Grant No. 2024-MSBA-42, in part by the Key Research and Development Program of Liaoning Province under Grant No. 2023JH26/10200011, and in part by the Fundamental Research Funds for the Central Universities under Grant No. N25YJS002.
\textit{(Corresponding Author: Gaochang Wu.)}}
\thanks{Mingxiu Cai, Zhe Zhang, Gaochang Wu, and Tianyou Chai are with the State Key Laboratory of Synthetical Automation for Process Industries, Northeastern University, Shenyang, China. Email: 2410285@stu.neu.edu.cn, zhangzhe17@stumail.neu.edu.cn, \{wugc, tychai\}@mail.neu.edu.cn.}
}

\markboth{Journal of \LaTeX\ Class Files,~Vol.~14, No.~8, August~2021}%
{Shell \MakeLowercase{\textit{et al.}}: A Sample Article Using IEEEtran.cls for IEEE Journals}


\maketitle

\begin{abstract}
The remarkable success of reconstruction-based methods in Unsupervised Anomaly Detection (UAD) lies in their ability to identify and localize anomalies by modeling discrepancies between input images and their reconstructed counterparts. However, these approaches often struggle to capture subtle anomalies and tend to produce blurred anomaly boundaries, which significantly limits their effectiveness, particularly in complex multi-class scenarios. To address these issues, we present XMatchAD, a novel UAD framework that reinterprets the task from a pseudo cross-modal matching perspective. Specifically, the input and reconstructed images are treated as two complementary modalities and their matching relationships are precisely exploited for anomaly detection. First, a pre-trained feature extractor is employed to encode discriminative representations. Second, an attention-guided cross-modal matching mechanism is introduced to match local inter-modal anomaly-related patterns while mutually refining the features. This enhances the sensitivity to anomalies with diverse shapes and subtle deviations and significantly improves the precision of anomaly detection and localization. Third, we design an adaptive frequency-aware fusion module that further delineates sharp anomaly boundaries through the coupling of high-frequency components from cross-modal multi-scale representations. Comprehensive evaluations on MVTec-AD, VisA, and MPDD benchmarks demonstrate that our method consistently achieves superior performance, outperforming state-of-the-art methods in multi-class anomaly detection and localization. The code will be released at \href{https://github.com/Mingxiu-Cai/XMatchAD}{https://github.com/Mingxiu-Cai/XMatchAD}.
\end{abstract}


\section{Introduction}
\IEEEPARstart{A}{nomaly} Detection (AD) is a critical task that aims to identify and localize patterns that deviate significantly from normal behavior with little or no prior knowledge. It has been widely applied across various domains, including video surveillance \cite{ref1}, industrial inspection \cite{ref2}, and medical image analysis \cite{ref3}. However, the diversity of anomaly patterns (e.g., structural anomalies, surface anomalies), combined with the rarity of anomalous instances and the labor-intensive nature of manual labeling, poses significant challenges to obtaining a sufficient number of well-annotated abnormal samples, particularly in complex and dynamic scenarios. Consequently, Unsupervised Anomaly Detection (UAD) has attracted considerable attention for its ability to learn exclusively from normal data and detect deviations from the learned distribution \cite{ref4} \cite{ref5}, with recent efforts increasingly focused on developing unified models capable of handling multi-class \cite{ref6, ref7, ref8}. 

\begin{figure}[t!]
    \centering
    \includegraphics[width=0.99\linewidth]{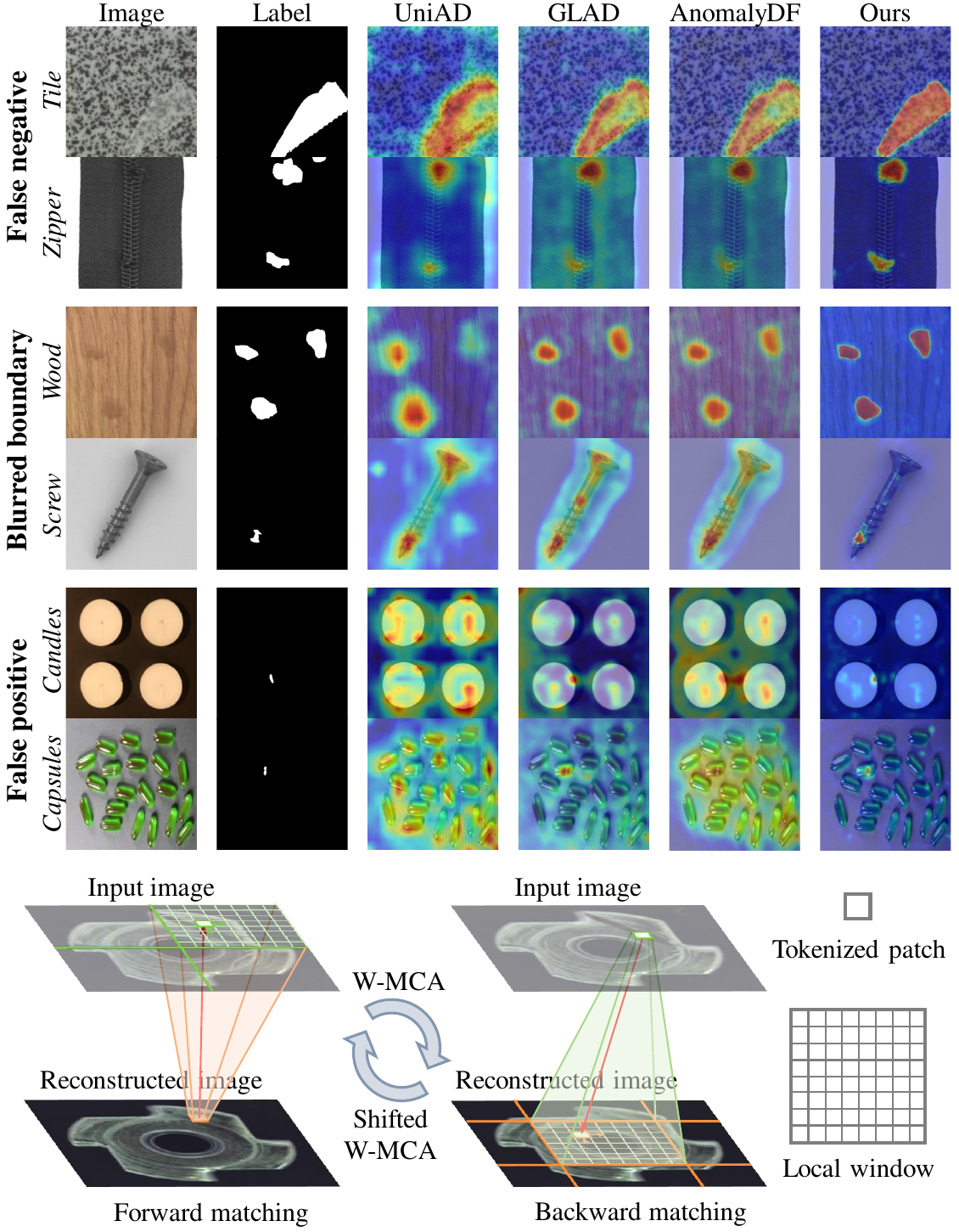} 
    \caption{Visualization of the predicted anomaly heatmaps from SOTA approaches and our method on MVTec-AD and VisA. Existing methods often encounter issues such as false negatives, false positives, and blurred boundaries. In contrast to these baselines, our method reconceptualizes anomaly detection as a \textbf{local matching} task via Window-based Multi-head \textbf{Cross-Attention} (W-MCA), leading to more precise and reliable localization of anomalies with sharper boundaries.}
    \label{fig:fig1}
    \vspace{-4mm}
\end{figure}

Existing UAD methods can be broadly divided into feature embedding-based and reconstruction-based paradigms. The former learns or pre-trains a latent space that captures the intrinsic characteristics of normal data. During inference, input samples are compared against the learned distribution to identify anomalies \cite{ref9} \cite{ref10}. However, when the distribution of test data differs from that used during pre-training, biased feature representations may emerge, leading to performance degradation. Reconstruction-based UAD methods, on the other hand, model the distribution of normal patterns using architectures such as AutoEncoders (AEs) \cite{ref11}, Variational AutoEncoders (VAEs) \cite{ref12}, and Generative Adversarial Networks (GANs) \cite{ref13}. They attempt to reconstruct anomalies and compute pixel-wise differences between the input and its anomaly-free reconstruction to detect anomalies. However, due to their limited reconstruction capacity, such models often tend to replicate the input \cite{ref_shortcut_input}, regardless of whether it contains defects, rather than truly modeling the joint distribution of normal data \cite{ref14} \cite{ref15}. To address these limitations, a self-supervised paradigm is adopted to generate normal-like reconstructions from anomalous inputs within diffusion-based models \cite{ref_25, ref30, ref31}, under the constraint of high-quality reconstruction and robust density estimation. This mechanism effectively alleviates the ``identical shortcut'' phenomenon \cite{ref_shortcut} and reduces ambiguous reconstructions, demonstrating promising performance in anomaly detection tasks.

However, most reconstruction-based methods suffer from three critical issues: detection uncertainty, insufficient sensitivity to subtle anomalies, and blurred anomaly boundaries, which in turn diminish overall detection performance and reliability. Firstly, existing reconstruction-based models \cite{ref18, ref19, ref21} struggle to clearly separate the distributions of normal and abnormal regions, which inevitably results in detection uncertainty (false negatives or false positives). 
Second, existing methods \cite{ref30, ref25-mdps,ref25-ddad,ref25-uniad} rely on parameter-free similarity measures between inputs and reconstructions. While these approaches exhibit strong responses to large or visually salient anomalies, they often lack the sensitivity required to detect small-scale or subtle anomalies \cite{ref23} \cite{ref24}. Moreover, the global similarity comparison strategy tends to introduce noise, increasing the likelihood of misclassifying normal regions as anomalous, as illustrated in Fig.~\ref{fig:fig1}. Third, existing models struggle to accurately delineate anomaly boundaries, often producing blurred predictions that obscure critical structural details, as evident in Fig.~\ref{fig:fig1}.

Motivated by the aforementioned observations, we reconceptualize reconstruction-based UAD as a pseudo cross-modal matching problem. In this formulation, potential anomalies are highlighted by explicitly contrasting and matching relationships between two modalized inputs: the original input samples and their reconstructed counterparts\footnote{Here, ``pseudo cross-modal matching'' refers to the comparison between inputs and reconstructions of the same data modality, unlike typical cross-modal matching (e.g., between images and text or audio).}. To this end, we present XMatchAD, a novel framework that is characterized by capturing inter-modal discrepancies to facilitate accurate localization of anomalous regions that deviate in appearance, structure, or both. The effectiveness of the proposed method stems from two key characteristics: 1) It explicitly enhances the discriminability between normal and anomalous samples, improving sensitivity to subtle abnormal signals while reducing detection uncertainty; 2) It incorporates high-frequency details and a local matching mechanism, enabling the model to delineate sharp anomaly boundaries.

Specifically, a pre-trained feature extractor is employed to capture salient features unique to each modality. This step serves to suppress irrelevant or noisy information while reducing computational complexity. The extracted features are then fed into the attention-guided cross-modal matching module, where intra-modal interactions are first performed via self-attention to refine representations and yield deep semantics-rich features. Subsequently, inter-modal feature matching is conducted via cross-attention to model correspondences between modalities. This matching process achieves comprehensive inter-modal comparison, allows the model to enhance anomaly detection capability with various shapes, and mitigates detection uncertainty. Given that anomalies often manifest in spatially continuous and localized regions, exploring matching correlations between all feature tokens introduces unnecessary computational overhead and inefficiency. Thus, a shift-window matching mechanism (Fig.~\ref{fig:fig1}, bottom) is adopted to focus on local regions with diverse locations and spatial scales. This approach aids in constructing local contextual dependencies, contributing to the detection of subtle anomalies and the accurate delineation of anomaly boundaries. Ultimately, an adaptive frequency-aware fusion module is introduced, with a particular emphasis on the high-frequency detail information inherent within feature maps. Inspired by the traditional high-pass filtering approach~\cite{ref27}, this mechanism further enhances segmentation accuracy and facilitates the detection of anomalies with diverse orientations and structures.

We summarize the contributions as follows:
\begin{itemize}
\item We propose XMatchAD, a novel method for multi-class anomaly detection. We reinterpret reconstruction-based UAD as a pseudo cross-modal matching process, where the input image and its reconstructed counterpart are treated as two pseudo modalities. Under this perspective, anomaly score maps are generated through local matching using the window-based cross-attention mechanism.

\item We propose an adaptive frequency-aware fusion module, which effectively captures high-frequency components within feature maps to enhance boundary accuracy and detection performance for anomalies with varying orientations and structures.

\item Our method demonstrates strong performance in multi-class anomaly detection and location, as verified by extensive experiments, achieving state-of-the-art performance with 98.7, 92.8, and 94.6 (image-AUROC), 98.2/68.9, 98.6/42.1, and 99.1/48.9 (pixel-AUROC/pixel-AP) for detection and location on MVTec-AD, VisA, and MPDD datasets, respectively.
\end{itemize}

\section{Related works}
\subsection{Reconstruction-based Anomaly Detection}
Reconstruction-based methods operate under the hypothesis that models can accurately exploit the feature distribution of normal samples and faithfully recover them, while producing significant deviations when presented with anomalous inputs. Methods based on AEs and VAEs are widely applied in UAD. However, due to their strong generalization capabilities, these models often exhibit reconstruction ambiguity and equally powerful reconstruction of anomalies, inflicting erroneous anomaly predictions. Methods built upon GANs detect anomalies by comparing the input image with random samples generated from the latent space through a discriminative network. For example, SCADN \cite{ref29} leverages multi-scale striped masks to reconstruct missing regions in normal samples, thereby learning rich semantic context and enabling effective UAD through reconstruction errors. Nevertheless, GAN-based methods frequently encounter training instability and are susceptible to mode collapse, producing reconstructions that lack diversity and occasionally exhibit semantic inconsistency or vacuity. 

Recently, diffusion-based anomaly detection methods have garnered increasing attention due to their powerful reconstruction capabilities. DiffAD \cite{ref_25} introduces latent diffusion augmented with noisy condition embedding and interpolated channels to tackle the challenge of low reconstruction quality. DiffusionAD \cite{ref19} enhances both reconstruction accuracy and realism through a noise-to-norm paradigm, featuring a rapid one-step denoising process and norm-guided improvements. GLAD \cite{ref30} dynamically predicts specific denoising steps for each sample and employs synthetic abnormal instances to cope with varying reconstruction complexities and noise characteristics, achieving flexible and precise anomaly-free reconstructions. DiAD \cite{ref31} further advances this direction by integrating diffusion-based generation with semantic-aware guidance and multi-scale feature analysis to accurately identify and localize multi-class anomalies.

Previous methods that attempt to model normal and anomalous samples within a unified distribution often struggle to capture potential anomaly patterns effectively. This limitation results in vague detection outcomes and ambiguous anomaly boundaries, particularly under multi-class scenarios. To address these challenges, we reconceptualize reconstruction-based anomaly detection as a pseudo cross-modal learning task. This formulation integrates unimodal representation learning with inter-modal interactions, thereby enhancing the model's discriminative capacity through more structured and effective matching mechanisms.

\subsection{Cross-Modal Learning}
Cross-modal learning exploits the interactions among different modalities that often encapsulate complementary information. This paradigm underscores the flexibility of handling heterogeneous data and bridges the gap across distinct domains, enabling models to achieve a more holistic understanding and improved decision-making performance, which has thus been widely adopted in autonomous driving \cite{ref32, ref33, ref34}, video question answering \cite{ref35}, vision-and-language navigation \cite{ref36} \cite{ref37}, medical data analysis \cite{ref38}, and visual-text retrieval \cite{ref39}.

A common strategy in cross-modal learning is to train modality-specific learners to independently capture informative representations, which are then refined through inter-branch fusion to boost the model's overall capability. For example, Ding et al. \cite{ref40} proposed a region-aware fusion method with probability maps that helps to partition the features extracted from four modalities individually into distinct regions. A modality-wise attention mechanism is then performed for region-level fusion for adaptive and effective cross-modal tumor segmentation. Zhao et al. \cite{ref41} feed visible and infrared images into a U-Net \cite{ref_unet} fusion module composed of Restormer \cite{ref42} and CNN blocks to capture cross-modal dependencies. Motivated by the inherent invariance of imaging responses to transformations, the generated pseudo-sensing images should be aligned with fused outputs to boost the representation quality. In a related approach, Li et al. \cite{ref43} employ sparse depth supervision and LiDAR occupancy guidance to enhance representations. An attention-based fusion module is then implemented to aggregate modality-aware features for prediction. Furthermore, Li et al. \cite{ref44} develop an adaptive cross-modal fusion mechanism that selectively integrates modality-specific features based on their contextual relevance.

Another alternative scheme in cross-modal learning involves constructing a shared latent space where features from different modalities are projected for direct comparison, alignment, or transformation. For example, Jiang et al. \cite{ref45} embed image and text features by a dual-stream architecture, and further propose similarity distribution matching to effectively amplify the discrepancy between negative pairs while reinforcing the correlation between positive pairs. To bridge the semantic-visual gap, Lu et al. \cite{ref46} design a semantic-guided module that adaptively selects relevant visual information from dual visual modalities, coupled with a semantic-visual alignment strategy to improve generalization for unseen classes. In a similar vein, Li et al. \cite{ref47} construct a framework that jointly optimizes the alignment of unimodal representations, masked language modeling, and image-text correspondence via contrastive learning.

Inspired by these advancements, we rethink original inputs and corresponding reconstructions generated by the diffusion model as distinct modalities, concentrating on different visual characteristics of anomalies through inter-modal information matching and interaction. We stack self-attention and cross-attention to simultaneously encode unimodal intrinsic features and investigate inter-modal mismatches to enhance the discriminative nature of the model.

\begin{figure*}[t]
    \centering
    \includegraphics[width=0.99\textwidth]{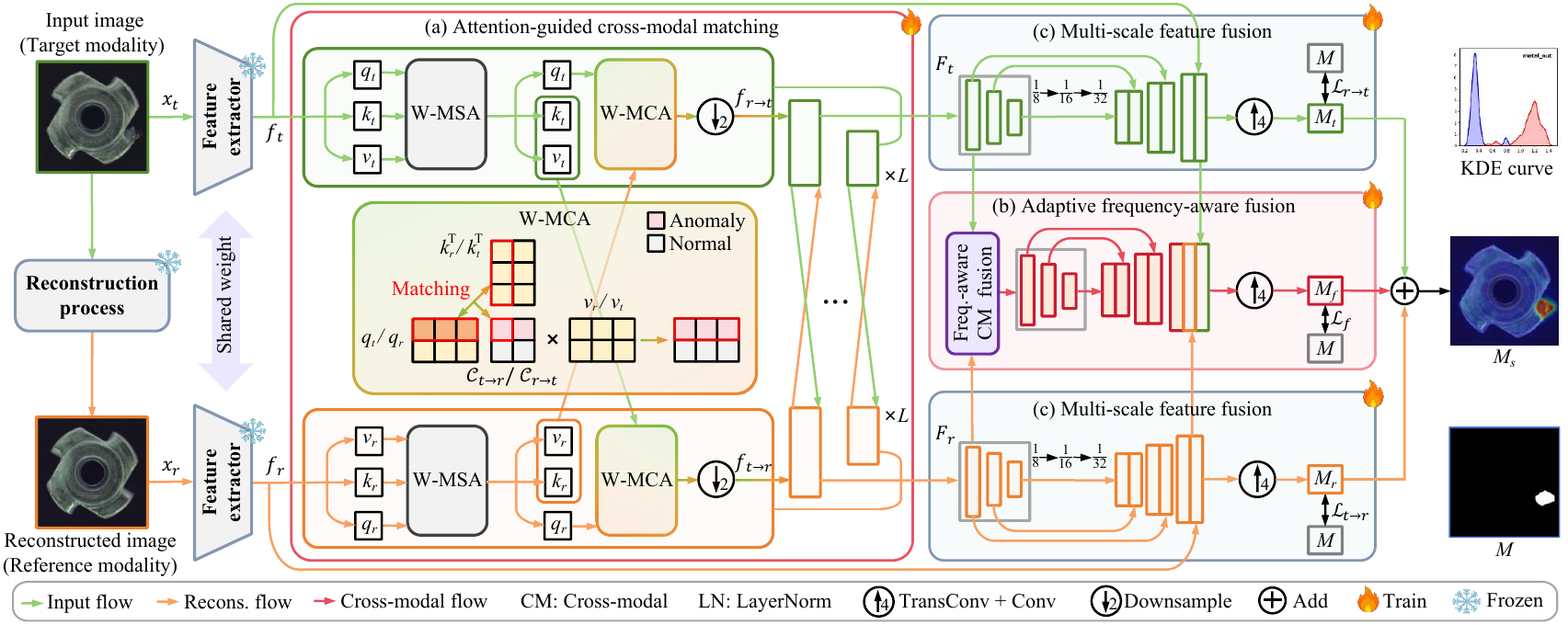}
    \caption{Overview of the proposed XMatchAD. We reformulate UAD as a pseudo cross-modal matching task, where the input and its reconstruction are treated as distinct modalities. Anomalies are identified by measuring consistency and inconsistency between them. (a) An attention-guided cross-modal matching module is introduced, leveraging W-MSA (Window-based Multi-head Self-Attention) and W-MCA (Window-based Multi-head Cross-Attention) to enhance feature representations and enable fine-grained matching. Before that, a pre-trained encoder first extracts intra-modal features from each modality. (b) An adaptive frequency-aware fusion module sharpens anomaly boundaries by coupling high-frequency components across multi-scale features from both modalities, producing an intermediate anomaly map $M_{f}$. (c) Multi-scale fusion is further applied to generate two additional maps $M_t$ and $M_r$ from each modality. The final anomaly prediction is obtained by averaging $M_f$, $M_t$, and $M_r$.}
    \label{fig:fig2}
    \vspace{-4mm}
\end{figure*}

\section{Methodology}
Some reconstruction-based UAD methods simulate anomaly generation to compensate for the scarcity of abnormal samples during training. These methods attempt to reconstruct the normal visual appearance and subsequently compare it with the input to determine where anomalies are present. However, existing works face several challenges: 1) Insufficient matching: the poor ability to correctly distinguish fine-grained and near-in-distribution anomalies. 2) Blurred boundary delineation, as illustrated in Fig.~\ref{fig:fig1}. Defeating these limitations requires designing more precise and robust discriminant architectures capable of handling complex industrial scenarios. Motivated by this, we propose a novel pipeline, XMatchAD, as depicted in Fig.~\ref{fig:fig2}. The framework comprises three key components: attention-guided cross-modal matching, adaptive frequency-aware fusion, and anomaly score generation.

\subsection{Problem Formulation}

Given an input sample $x_t \in \mathbb{R}^{C \times H \times W}$, reconstruction-based methods, e.g., DRAEM \cite{ref18} using a U-Net and GLAD \cite{ref30} using a diffusion model, learn the entire distribution of normal visual appearance using only normal samples, and produce the reconstructed counterpart $x_r \in \mathbb{R}^{C \times H \times W}$, where $C$, $H$, and $W$ are the number of channels, height, and width, respectively. Anomaly detection is then performed by either modeling normal and abnormal samples within a joint feature distribution or by performing feature-level matching (e.g., cosine similarity) between $x_t$ and $x_r$ to identify anomalies:
\begin{equation}\label{eq:cos_sim}
\text{sim}(f_t, f_r)=\frac{f_tf_r^T}{||f_t||\cdot||f_r||},
\end{equation}
where $\text{sim}(\cdot, \cdot)$ indicates cosine similarity, and $f_t$ and $f_r$ are the encoded features extracted by a pre-trained encoder (such as DINOv2 \cite{ref48}). However, the absence of dedicated components for clearer comparative reasoning inherently limits the model’s ability to finely capture discriminative characteristics, increasing the risk of ambiguity or confusion in anomaly segmentation, as illustrated in Fig.~\ref{fig:fig1}. This issue is further pronounced in multi-class scenarios, where subtle and class-specific anomalies are more likely to be neglected, indicating the need for more precise and powerful comparison and matching mechanisms.

Motivated by recent advances in cross-modal learning, we reformulate the reconstruction-based anomaly detection (Eqn.~(\ref{eq:cos_sim})) as a pseudo cross-modal matching task, wherein a learnable matching process is implemented via a cross-attention mechanism. In this formulation, we treat the input samples and their corresponding anomaly-free reconstructions as two distinct modalities and design a model to process information from the two modality branches separately, denoted as $\text{Att}(f_t, f_r)$, where $t$ refers to the target modality (input sample) and $r$ denotes the reference modality (reconstructed counterpart). Unlike methods tailored to specific anomaly types, the proposed pseudo-modality formulation enables mutual guidance between normal and abnormal cues, thereby enhancing both the matching precision and discriminative capacity of the network. This formulation encourages the learning of more robust and expressive representations that explicitly disentangle anomalies within the latent space. As a result, the model exhibits heightened sensitivity to subtle or structurally irregular deviations, effectively reducing detection uncertainty.

Following prior works, we perform feature-level comparison and matching. Specifically, a pre-trained DINOv2 \cite{ref48} is used to extract discriminative features from the reference ($x_r$) and target ($x_t$) modalities. The resulting last-layer features, $f_r, f_t \in \mathbb{R}^{C' \times H' \times W'}$, are used for subsequent processing.

\subsection{Attention-Guided Cross-Modal Matching}
The attention-guided cross-modal matching module, which consists of intra-modal feature enhancement and attention-guided cross-modal feature matching, as shown in Fig.~\ref{fig:fig2} (a). It serves to suppress matching noise and improve detection accuracy by reducing boundary ambiguity and enhancing sensitivity to subtle anomalies.

\subsubsection{Intra-modal feature enhancement} \label{sec:MSA}
Note that the sole reliance on $f_r$ and $f_t$ forces the model to handle the entire image uniformly, limiting its ability to effectively highlight important local details and salient regions. Consequently, we incorporate a self-attention mechanism, as illustrated by W-MSA in Fig. \ref{fig:fig2} (a), to assist the model in dynamically adjusting the importance of features at different spatial locations, thereby enhancing contextual understanding and enriching the semantic representation. Given that anomalies are typically manifested in localized regions and characterized by uneven distributions, it suggests that the window-based Swin Transformer (Swin-T) \cite{ref26} is more suitable for anomaly detection tasks than the vanilla Transformer \cite{ref53}. Swin-T conducts multi-head self-attention within windows at multiple scales, allowing for dynamic weighting of local features and effective perception of subtle details. Accordingly, $f_r$ and $f_t$ are fed separately into two Swin-T branches, where they are projected into query $q$, key $k$, and value $v$ representations, with each feature point treated as a token:
\begin{equation}
\begin{aligned}
\label{eqn:eqn1}
q_n, k_n, v_n &= \text{Proj}(f_n), \ \text{Proj}(f_n),  \ \text{Proj}(f_n), \\
\text{Proj}(a) &= \text{Linear}\big(\text{WP}\big( \text{LN}(a)\big)\big),
\end{aligned}
\end{equation}
where $n$ represents $r$ or $t$, Proj indicates the projection operation, LN refers to layer normalization, Linear denotes the linear mapping layer, and WP is the window partitioning operation, including regular window partitioning with equally sized windows and shifted window partitioning with varying window sizes. The refined feature $f_n$ is then computed through the self-attention as follows:
\begin{equation}
f_n \leftarrow \text{Softmax}\left(\frac{q_n k_n^T}{\sqrt{d}}\right) v_n + f_n,
\end{equation}
where Softmax denotes the softmax function. $f_n$ provides comprehensive semantic information and embeds richer contextual dependencies, allowing the model to focus on subtle or localized details and improve performance. 

\subsubsection{Cross-modal feature matching}\label{sec:matching}
The core objective of attention-guided cross-modal feature matching is to establish fine-grained correspondences between modalities by leveraging their complementary cues to further refine feature representations. To this end, we propose a cross-modal matching mechanism, as illustrated by W-MCA in Fig. \ref{fig:fig2} (a), which facilitates bidirectional knowledge interaction, i.e., forward and backward matching, between the reference and the target in feature space. Specifically, we employ window-based cross-attention to implement this localized feature matching, which supports precise alignment of local patterns and subtle discrepancies while maintaining computational efficiency.

\textbf{Forward matching:} To enhance the reference modality feature $f_r$ using information from the target modality $f_t$, we reformulate the cosine similarity-based matching in Eqn.~(\ref{eq:cos_sim}) as an attention-guided matching process, defined as follows:
\begin{equation}
\mathcal{C}_{t \rightarrow r} = \text{Softmax}\left(\frac{q_r k_t^T}{\sqrt{d}}\right),
\end{equation}
where $f_r$ is projected as query $q_r$, while key $k_t$ and value $v_t$ are derived from $f_t$ by the Proj operation, and $\mathcal{C}_{t \rightarrow r}$ represents the normalized inter-modal matching correlation from the target modality to the reference modality. In this case, given the target modality as a contextual prompt, the matching correlation is used to selectively emphasize both notable discrepancies and shared semantic cues between modalities. Then, we further refine the reference feature through a weighted aggregation guided by the cross-modal matching correlation, followed by a non-linear transformation, as detailed below:
\begin{equation}
\begin{aligned}
f_{t \rightarrow r}' &= \mathcal{C}_{t \rightarrow r} v_t + f_r, \\
f_{t \rightarrow r} &= \text{MLP}(f_{t \rightarrow r}') + f_{t \rightarrow r}',
\end{aligned}
\end{equation}
where $f_{t \rightarrow r}'$ weights the feature by the matching correlation, MLP is a multi-layer perceptron, and $f_{t \rightarrow r}$ denotes the resulting refined feature representation generated through forward matching. 

\textbf{Backward matching:} Similarly, the backward matching operation is performed to update $f_t$ using $f_r$, that is,
\begin{equation}
\begin{aligned}
\mathcal{C}_{r \rightarrow t} &= \text{Softmax}\left(\frac{q_t k_r^T}{\sqrt{d}}\right),\\
f_{r \rightarrow t}' &= \mathcal{C}_{r \rightarrow t} v_r + f_t, \\
f_{r \rightarrow t} &= \text{MLP}(f_{r \rightarrow t}') + f_{r \rightarrow t}',
\end{aligned}
\end{equation}
where $\mathcal{C}_{r \rightarrow t}$ stands for the normalized inter-modal matching correlation from the reference modality to the target modality, and $f_{r \rightarrow t}$ represents the corresponding refined feature representation obtained through backward matching.

Therefore, this bidirectional cross-attention matching alleviates potential information loss caused by directional asymmetry and establishes mutual contextual referencing between modalities. It not only enables mutual enhancement that preserves intra-modal semantics but also reinforces inter-modal consistency. Such a mechanism significantly improves sensitivity to anomalous deviations, thereby reducing detection uncertainty and achieving a more precise delineation of anomaly boundaries. 

As the refined features propagate through $L$ layers of attention-guided cross-modal matching with downsampling, multi-scale features are produced as follows: 
\begin{equation}
\begin{aligned}
F_r = \{ f_{t \rightarrow r}^l\}_0^{L-1}, F_t = \{ f_{r \rightarrow t}^l\}_0^{L-1},
\end{aligned}
\end{equation}
where $f_{t \rightarrow r}^l\in\mathbb{R}^{(C' \cdot 2^l) \times \frac{H'}{2^l} \times \frac{W'}{2^l}}$ and $f_{r \rightarrow t}^l\in \mathbb{R}^{(C' \cdot 2^l) \times \frac{H'}{2^l} \times \frac{W'}{2^l}}$ represent the bidirectional features at $l$-th scale level, and $F_r$ and $F_t$ are the resulting bidirectional feature sets. These multi-scale representations refined through cross-modal matching are subsequently fed into the adaptive frequency-aware fusion (see Sect. \ref{sec:Fusion}) and multi-scale feature fusion (see Sect. \ref{sec:anomaly_score}) for the final anomaly inference.

\subsection{Adaptive Frequency-aware Feature Fusion}\label{sec:Fusion}
\begin{figure*}[ht]
    \centering
    \includegraphics[width=0.95\textwidth]{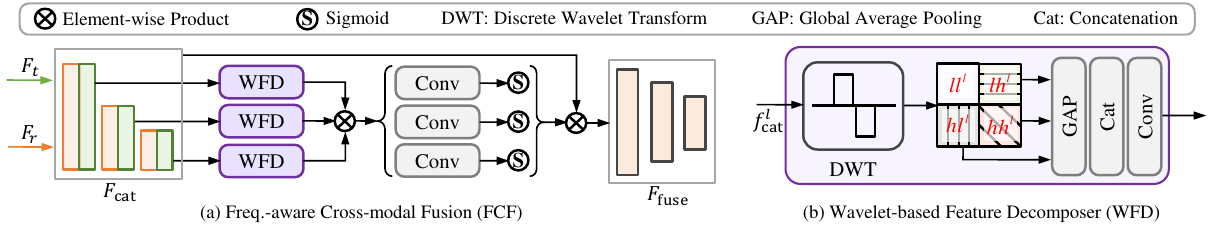}
    \vspace{-1mm}
    \caption{Design of the adaptive frequency-aware fusion. (a) Frequency-aware Cross-modal Fusion (FCF). (b) Wavelet-based Feature Decomposer (WFD).}
    \label{fig:fig3}
    \vspace{-4mm}
\end{figure*}

Effective anomaly segmentation relies on the synergistic integration of spatial details and semantic knowledge. Moreover, multi-scale feature fusion encourages the model to make sense of both small-scale textures and large-scale patterns, creating a more informative and diverse feature space that supports precise anomaly detection. Therefore, we design a Frequency-aware Cross-modal Fusion (FCF) component as shown in Fig.~\ref{fig:fig3}, where spatial features are decomposed into distinct frequency subbands with rich details to collaboratively guide the fusion of features from dual modalities for accurate boundary delineation. Then, multi-scale feature fusion is applied to the fused features to generate prediction, as shown in Fig.~\ref{fig:fig2} (b).

\textbf{Frequency-aware cross-modal fusion.} Frequency domain features possess unique advantages, notably their capacity to capture fine-grained details like edges and textures, which are often altered in anomalous regions. They also provide a complementary representation by emphasizing local variations and high-frequency cues critical for detecting subtle anomalies. Motivated by these properties, we incorporate frequency-domain information into spatial features to enhance the model's discriminative capability. Owing to its strong capabilities in both frequency and spatial representation, the wavelet transform is adopted to alleviate blurred boundaries. FCF builds on high-pass filtering methods by leveraging the concept that spatial details are predominantly concentrated in the high-frequency spectrum \cite{ref_freq}. The implementation of FCF involves two key steps: decomposing spatial features into frequency information and coupling multi-scale high-frequency details to guide the fusion of spatial features.

Initially, FCF merges the bidirectional feature sets $F_r$ and $F_t$ along the channel dimension to yield a new enriched feature set $F_\text{cat}$:
\begin{equation}
F_\text{cat} = \{ f^l_\text{cat}\}_0^{L-1},\ \ f^l_\text{cat}=\text{Cat}(f_{t \rightarrow r}^l, f_{r \rightarrow t}^l),
\end{equation}
where Cat denotes concatenation along the channel dimension.
These features are first fed into the wavelet-based feature decomposition, as illustrated in Fig. \ref{fig:fig3} (b), to generate $L$ frequency domain features. Unlike the Fourier transform, the wavelet transform provides a more efficient mechanism for decomposing images into both low-frequency semantic structures and high-frequency detail features. Here, Discrete Wavelet Transform (DWT) \cite{ref28} is used to decompose mixed-scale spatial features into one low-frequency subband $ll^l$ and three direction-sensitive high-frequency subbands $\{lh^l, hl^l, hh^l\}$ of the same dimension:
\begin{equation}
\begin{aligned}
    & \{ll^l, \ lh^l, \ hl^l,\ hh^l\} = \text{DWT}(f^l_\text{cat}), \\
    &ll^l, lh^l, hl^l, hh^l \in \mathbb{R}^{(C'  \cdot 2^l) \times \frac{H'}{2^{l+1}} \times \frac{W'}{2^{l+1}}}.
    \label{eq:hl}
\end{aligned}
\end{equation}
High-frequency information can efficiently preserve edges, textures, and local variations, which tend to remain consistent in normal cases, but may reveal irregular or abrupt changes in anomalous ones. Focusing on these high-frequency cues, the model further achieves improved anomaly segmentation accuracy. Therefore, we derive the compact transformed representation $\hat{h}^l$ using high-frequency components for each stage of features via
\begin{equation}
    \hat{h}^l = \text{Cat}\big( \text{GAP}(lh^l),  \text{GAP}(hl^l),  \text{GAP}(hh^l) \big),\ \hat{h}^l \in \mathbb{R}^{(3 \cdot C'\cdot 2^l) \times 1},
\end{equation}
where GAP stands for global average pooling.

Afterwards, we make full use of the high-frequency details from different stages to guide the fusion of spatial features. We embed $\{\hat{h}^0,..,\hat{h}^{L-1}\}$ into a group of parallel convolutional layers to align $\hat h^l$ in a unified dimension $C'\times 1$, and an element-wise multiplication is performed to tightly couple high-frequency representations across levels:
\begin{equation}
    h = \prod_{l=0}^{L-1} \text{Conv}(\hat{h}^l).
    \label{eq:conv}
\end{equation}
As illustrated in Fig. \ref{fig:fig3} (a), inter-branch coupling enforces mutual dependency, ensuring meaningful and robust representation learning, where failure in one branch will severely degrade the overall outcome. The $h$ is transformed back to the original shape using a Sigmoid-activated convolution, producing adaptive weights that attempt to identify the optimal combination of fused cross-modal features while preserving critical spatial information:
\begin{equation}
\begin{aligned}
    f^l_{\text{fuse}} &= f^l_\text{cat} \odot \big(\text{Sigmoid} \big(\text{Conv}(h)\big)\big), \\ 
    F_{\text{fuse}} &= \{ f_{\text{fuse}}^l \}_0^{L-1},
    \label{eq:Gl}
\end{aligned}
\end{equation}
where $\odot$ denotes element-wise multiplication, and $F_{\text{fuse}}$ is the resulting fused feature set.

\textbf{Multi-scale feature fusion.}\label{Multi-scale_feature_fusion} We conduct a U-Net-like multi-scale fusion on the feature set $F_{\text{fuse}}$, as illustrated in Fig.~\ref{fig:fig2} (b). This hierarchical fusion strategy aggregates multi-level contextual information by progressively combining high-resolution spatial details with high-level semantic features through a series of upsampling operations and lateral connections. To facilitate more effective gradient propagation and preserve spatial fidelity, we incorporate skip connections from earlier layers into the feature stream, ultimately producing an anomaly score map $M_{f}$.

\subsection{Anomaly Score Generation}\label{sec:anomaly_score}
Similarly, multi-scale feature fusion, as described in Section~\ref{sec:Fusion}, is independently employed in the two branches corresponding to the reference modality and target modality feature sets to generate  corresponding predictions $M_r$ and $M_t$, as illustrated in Fig.~\ref{fig:fig2} (c).

For anomaly localization, the outputs $M_{f}$, $M_r$, and $M_t$ from the three branches are normalized respectively using the softmax function and then averaged to generate a consolidated anomaly prediction $M_s$:
\begin{equation}
M_s = \text{Avg}\big(\text{Softmax}(M_{f}, M_r, M_t)\big),
\label{eqn:eqn_Ms}
\end{equation}
where Avg represents the average operation and Softmax denotes the softmax activation function. As for single-head segmentation models, the multi-head architecture offers powerful flexibility by allowing parallel learning of feature representations from different branches during training and inference.

By learning complementary information, each head contributes to more comprehensive feature representations. For anomaly detection, the image-level anomaly score is derived by averaging the top $K$ pixel values from the final anomaly score map $M_s$ with $K$ empirically set to 250.

\begin{table*}[t]
\renewcommand{\arraystretch}{0.85}
    \centering
    \caption{Multi-class anomaly detection and location performance on three datasets with six metrics. Best results are highlighted in \textbf{bold}.} 
    \label{tab:tab_multi_datasets}
    \vspace{-2mm}
    \footnotesize
\setlength{\tabcolsep}{11pt}
    \begin{tabular}{ll c c c c c c c }
        \toprule
        \multirow{2}{*}{Dataset} & \multirow{2}{*}{Method} & \multirow{2}{*}{Input size}  & \multicolumn{3}{c}{Image-level} & \multicolumn{3}{c}{Pixel-level} \\
\cmidrule(lr){4-6} \cmidrule(lr){7-9}
&  &  & AUROC & AP & F1-max & AUROC & AP & F1-max \\
\midrule
\multirow{11.5}{*}{MVTec-AD} & DRAEM \cite{ref18} & \multirow{10}{*}{$256^2$} &88.1 & 94.7 & 92.0  & 87.2 & 52.5 & 48.6\\
& UniAD \cite{ref25-uniad} && 96.5 & 98.8 & 96.2 & 96.8 & 43.4 &49.5\\
   & SimpleNet\cite{ref49} && 95.3 & 98.4 & 95.8 & 96.9 & 45.9 & 49.7\\
   & DeSTSeg \cite{ref23} && 89.2 & 95.5 & 91.6& 93.1 & 54.3 & 50.9\\
   &DiAD\cite{ref31}  && 97.2 &99.0 &96.5& 96.8 & 52.6 & 55.5\\
   &GLAD\cite{ref30} && 97.5 &98.8 &96.8 & 97.3 & 58.8 & 59.7\\
   &ViTAD\cite{ref50} && 98.3 & \textbf{99.4} &97.3 & 97.7 & 55.3 &58.7\\
   &AnomalDF\cite{ref51}  &&96.8 & 98.6 &97.1& 98.1 & 61.3 & 60.8\\
   &XMatchAD (Ours)  && \textbf{98.4} & \textbf{99.4} & \textbf{97.4} & \textbf{98.2} & \textbf{69.8} & \textbf{66.3} \\
  \cmidrule{2-9}
   &Dinomaly\cite{dinomaly} & \multirow{2}{*}{$392^2$} & \textbf{99.6} & \textbf{99.8} &\textbf{99.0}& 98.4 & 69.3 & 69.2\\
   &XMatchAD$^*$ (Ours) &&\textbf{99.6} & \textbf{99.8} & \textbf{99.0} & \textbf{99.0} & \textbf{79.2} & \textbf{74.4}\\
   \midrule
   \multirow{11.5}{*}{VisA} & DRAEM \cite{ref18} & \multirow{10}{*}{$256^2$} &79.5 & 82.8 & 79.4 & 91.4 & 24.8 & 30.4\\
& UniAD \cite{ref25-uniad} && 85.5 & 85.5 & 84.4 & 95.9 & 21.0&27.0\\
   & SimpleNet\cite{ref49} && 87.2 & 87.0 & 81.8 & 96.8 & 34.7&37.8\\
   & DeSTSeg \cite{ref23}&& 88.9 & 89.0 & 85.2 & 96.1 & 39.6 & 43.4\\
   &DiAD\cite{ref31}  && 86.8 & 88.3 & 85.1 & 96.0 & 26.1 & 33.0\\
   &GLAD\cite{ref30} && 90.1 & 91.4 & 86.7 & 97.4 & 33.9 & 39.4\\
   &ViTAD\cite{ref50} && 90.5 & 91.7 &86.3 & 98.2 & 36.6 & 41.1\\
   &AnomalDF\cite{ref51}  && 90.5 & 91.4 & 86.2 & 97.4 & 39.6 & 40.4\\
   &XMatchAD (Ours)  && \textbf{94.2} & \textbf{95.5} & \textbf{89.8} & \textbf{99.0} & \textbf{46.5} & \textbf{48.8} \\
   \cmidrule{2-9}
   &Dinomaly\cite{dinomaly} & \multirow{2}{*}{$392^2$} & \textbf{98.7}& 98.9 & \textbf{96.2} & 98.7 & 53.2 & \textbf{55.7}\\
   &XMatchAD$^*$ (Ours) && \textbf{98.7} & \textbf{99.0} & 95.8 & \textbf{99.3} & \textbf{55.2} & 55.1 \\
   \midrule
   \multirow{10.5}{*}{MPDD} & DRAEM \cite{ref18} & \multirow{9}{*}{$256^2$} &88.3 &- & - & 88.3 & -& -\\
& UniAD \cite{ref25-uniad} && 87.5 & 83.2 & 85.1 & 95.6 & 19.0 & 25.6 \\
   & SimpleNet\cite{ref49} && 90.6 & 94.1 & 89.7 & 97.1 & 33.6 & 35.7\\
   & DeSTSeg \cite{ref23} && 92.6 &91.8 &92.8 &90.8 & 30.6 & 32.9\\
   &DiAD\cite{ref31}  && 85.8 & 89.2 & 86.5 & 91.4 & 15.3 & 19.2\\
   &GLAD\cite{ref30} && 90.8 & 90.5 & 90.2 & 98.0 & 40.0 & 40.6   \\
   &ViTAD\cite{ref50} && 87.4& 90.8 & 87.0 & 97.8 & 44.1 & 46.4\\
   &XMatchAD (Ours)  && \textbf{93.6} & \textbf{94.5} &\textbf{93.0} &\textbf{98.7}&	\textbf{46.2}&\textbf{46.9}
   \\
   \cmidrule{2-9}
   &Dinomaly\cite{dinomaly} & \multirow{2}{*}{$392^2$} & 97.2 & 98.4 & 96.0 & 99.1 & 59.5 & 59.4\\
   &XMatchAD$^*$ (Ours) &&\textbf{97.5} &\textbf{98.5}&\textbf{96.1}&\textbf{99.2}&	\textbf{60.2}&\textbf{60.2}\\
        \bottomrule
    \end{tabular}
    \vspace{-1mm}
\end{table*}

\subsection{Training and Inference}
We explore the pseudo cross-modal matching to product anomaly score maps $M_f$, $M_r$, and $M_t$, which are all expected to align closely with the ground truth mask $M$. Therefore, the overall loss can be expressed as follows:
\begin{equation}
\begin{aligned}
\mathcal{L}_{\text{total}} &=\alpha \cdot\mathcal{L}_f(M_{f}, M) + \beta \cdot \mathcal{L}_{t \rightarrow r}(M_r, M) \\ &+ \gamma \cdot \mathcal{L}_{r \rightarrow t}(M_t, M),
\end{aligned}
\end{equation}
where $\mathcal{L}_{t \rightarrow r}$, $\mathcal{L}_{r \rightarrow t}$, and $\mathcal{L}_f$ denote the BCE loss, $\alpha$, $\beta$, and $\gamma$ are tunable hyperparameters that balance the contributions of the individual prediction heads.

During the inference phase, the model produces three prediction outputs, and the anomaly score map is generated according to Eqn.~\eqref{eqn:eqn_Ms}, serving both pixel-level localization and image-level detection.

\section{Experiments}

\subsection{Experiments Setup and Implementation}
\subsubsection{Datasets} 
\textbf{MVTec-AD} is a benchmark dataset for industrial anomaly detection, comprising 15 categories (10 objects, 5 textures). It reflects realistic manufacturing conditions, with 3,629 normal training images and 1,725 test images, including normal and anomalous samples. \textbf{VisA} contains 8,659 training samples and 2,162 testing samples encompassing multiple defect types such as scratches, dents, spots, cracks, and structural damage. It covers 12 object categories featuring varied structures and spatial configurations, grouped into complex structures, multiple instances, and single instance types. \textbf{MPDD} is a dataset designed for surface defect detection in painted metal parts. It is composed of six categories that exemplify the typical challenges encountered in manual production environments. The dataset includes 888 normal samples for training and 458 samples for testing.

\subsubsection{Implementation Details}
We adopt the same self-supervised training strategy as commonly used in UAD methods~\cite{ref18,ref30,ref19}. For the reconstruction process, we employ GLAD~\cite{ref30} with 25 denoising steps to generate normal-like images. In this implementation, all input images are uniformly resized to a resolution of $256\times256$ to ensure consistency with baseline methods. A pre-trained DINOv2~\cite{ref48} serves as the feature extractor, utilizing the 12th-layer features as input to the downstream model. In addition, we consider a higher-resolution setting of $392\times392$, where Dinomaly~\cite{dinomaly} is used for the reconstruction process. This variant is denoted as XMatchAD$^*$, which adopts the fused multi-level DINOv2 features from Dinomaly as the inputs for both pseudo modalities.
The loss function comprises three sub-components, whose contributions are empirically balanced with weights $\alpha = 0.8$, $\beta = 0.1$, and $\gamma = 0.1$. $L$, $C'$, $H'$, and $W'$ are set to 3, 48, 28, and 28, respectively. 
The learning rate was set to $4 \times 10^{-3}$. 


\subsubsection{Evaluation metrics} 
Anomaly detection performance is evaluated using the Image-level Area Under the Receiver Operating Curve (I-AUROC), Average Precision (I-AP), and F1max (I-F1max), which quantify image-level anomaly accuracy. For pixel-level localization, we report Pixel-level AUROC (P-AUROC), Average Precision (P-AP), and F1max (P-F1max) to comprehensively assess segmentation performance.

\subsection{Comparison with SOTA Methods}
In this section, we present extensive qualitative and quantitative comparisons with State-Of-The-Art (SOTA) methods under multi-class UAD settings to validate the effectiveness and superiority of the proposed model. Besides the full model (XMatchAD), we also develop a lightweight version, XMatchAD$^*$, replacing the diffusion model with a lightweight reconstruction backbone Dinomaly \cite{dinomaly}. The compared baselines include embedding-based methods SimpleNet \cite{ref49}, DeSTSeg \cite{ref23}, AnomalDF \cite{ref51}, and reconstruction-based methods DRAEM \cite{ref18},  UniAD \cite{ref25-uniad}, DiAD \cite{ref31}, GLAD \cite{ref30}, Dinomaly\cite{dinomaly}, and ViTAD \cite{ref50}.

\subsubsection{Multi-class Anomaly detection}
The anomaly detection results evaluated on the \textbf{MVTec-AD} dataset are shown in Table~\ref{tab:tab_multi_datasets}. Our method achieves superior performance on various baselines spanning different architectural and methodological paradigms. For image-level anomaly detection, XMatchAD consistently outperforms DiAD, GLAD, and AnomalDF, achieving improvements of up to 0.6\%, 0.9\%, and 1.6\% in I-AUROC, respectively. 

This improvement stems from the cross-modal matching mechanism, which leverages local window shifts to provide a quasi-global perspective similar to vanilla cross-attention with substantially fewer parameter complexities. As a result, XMatchAD achieves more accurate image-level anomaly detection and improved sensitivity to subtle defects.


Compared to MVTec-AD, the \textbf{VisA} dataset poses a more realistic and challenging scenario, characterized by high intra-class variability, complex object structures, and subtle anomaly patterns. As shown in Table \ref{tab:tab_multi_datasets}, XMatchAD delivers absolute gains in I-AUROC/I-AP/I-F1max of 7.4\%/7.2\%/4.7\%, 4.1\%/7.2\%/3.1\%,  and 3.7\%/3.8\%/3.5\% over DiAD, GLAD, ViTAD. 

According to the results in Table~\ref{tab:tab_multi_datasets}, XMatchAD outperforms DiAD, GLAD, and ViTAD by 7.8\%/5.6\%/6.5\%, 2.8\%/4\%/2.8\%, and 6.2\%/3.7\%/6.0\% in I-AUROC/I-AP/I-F1max on the \textbf{MPDD} dataset. XMatchAD$^*$ also outperforms Dinomaly by 0.3\%/0.1\%/0.1\%, further validating its enhanced detection capability and robustness.


\begin{figure}[t]
    \centering
    \includegraphics[width=0.48\textwidth, page=1]{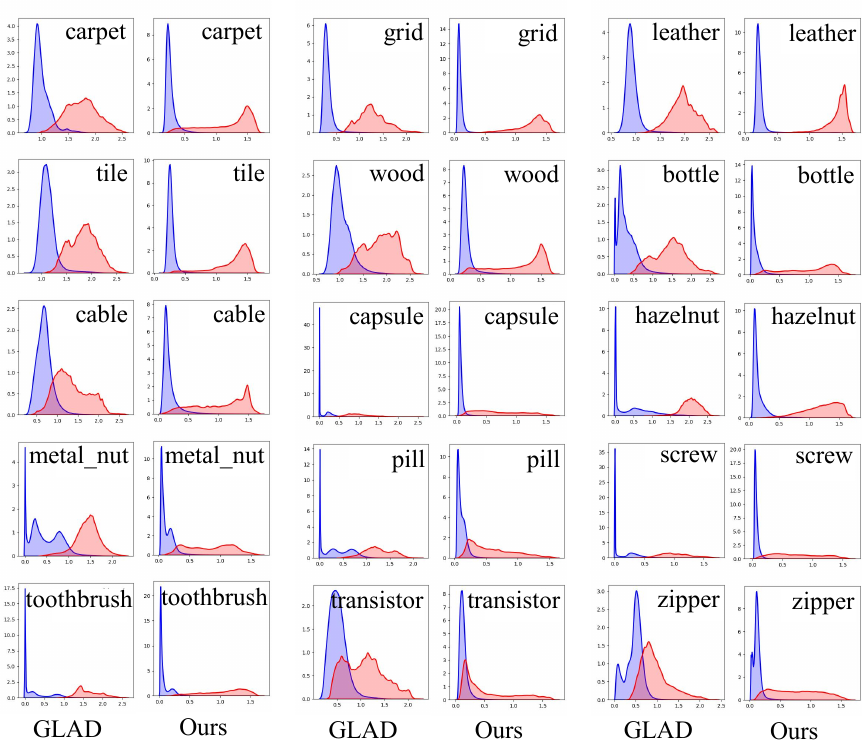} 
    \caption{Pixel-level anomaly score distributions for each category on MVTec-AD visualized through KDE curves. Pink denotes abnormal instances, whereas blue corresponds to normal ones.}
    \label{fig:fig_mvtec_kde_pixel}
    \vspace{-4mm}
\end{figure}

\begin{figure*}[t]
    \centering
    \includegraphics[width=\textwidth, page=1]{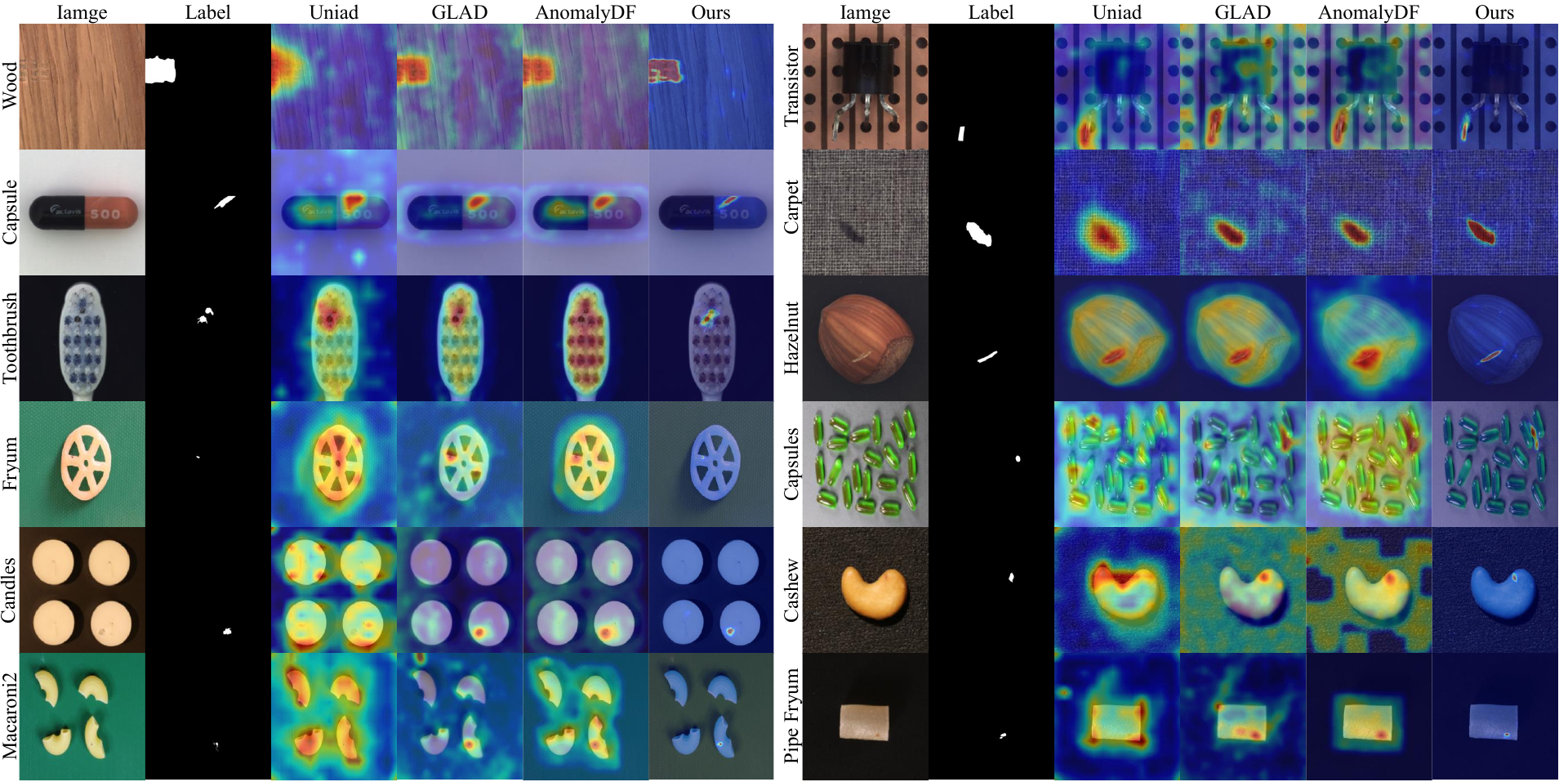} 
    \caption{Anomaly localization visualization on MVTec-AD and VisA datasets. From left to right are input images, ground truth, and heatmaps generated by UniAD, GLAD, AnomalyDF, and our method. }
    \label{fig:fig_Qualitative_Comparison}
    \vspace{-4mm}
\end{figure*}


\subsubsection{Multi-class Anomaly localization}
Unlike anomaly detection, which focuses on identifying whether a sample is anomalous, anomaly localization aims to precisely delineate the spatial regions that deviate from normal patterns. We perform a comprehensive evaluation of pixel-level metrics in three datasets. The average scores for anomaly localization on \textbf{MVTec-AD} are reported in Table~\ref{tab:tab_multi_datasets}. In terms of P-AUROC/P-AP/P-F1max, XMatchAD yields remarkable improvements of 1.4\%/17.2\%/10.8\%, 0.9\%/11.0\%/6.6\%, and 0.5\%/14.5\%/7.6\% over DiAD, GLAD, and ViTAD. Meanwhile, XMatchAD$^*$ also outperforms the baseline method Dinomaly by 0.6\%/9.9\%/5.2\%. Furthermore, the Kernel Density Estimation (KDE) \cite{ref52} curves in Fig.~\ref{fig:fig_mvtec_kde_pixel} reveal markedly reduced overlap in the pixel-level anomaly score distributions for our method relative to GLAD, reflecting enhanced discriminability between normal and anomalous instances. The reason for this lies in the fact that frequency-aware channel-wise weighting exploits high-frequency details to encourage sharper boundary delineation. On the \textbf{VisA} dataset, XMatchAD surpasses the baselines GLAD, ViTAD, and AnomalyDF by 1.6\%/12.6\%/9.4\%, 0.8\%/9.9\%/7.7\%, and 1.9\%/5.9\%/8.4\% in P-AUROC/P-AP/P-F1max, respectively. XMatchAD$^*$ outperforms Dinomaly by 0.6\%/2.0\% in P-AUROC/P-AP. Similarly, the results on \textbf{MPDD} in Table~\ref{tab:tab_multi_datasets} indicate that XMatchAD achieves improvements of 7.3\%/30.9\%/27.7\%, 0.7\%/6.2\%/6.3\% and 0.9\%/2.1\%/0.5\% in P-AUROC/P-AP/P-F1max over DiAD, GLAD and ViTAD. XMatchAD$^*$ XMatchAD$^*$ also outperforms Dinomaly by 0.1\%/0.7\%/0.6\%  These quantitative results highlight the compatibility and strong generalizability of our method in capturing diverse anomaly patterns. The image- and pixel-level KDE curves for VisA are presented in the Supplementary.

\subsubsection{Qualitative Comparison}
\begin{figure}[t]
    \centering
    \includegraphics[width=0.48\textwidth, page=1]{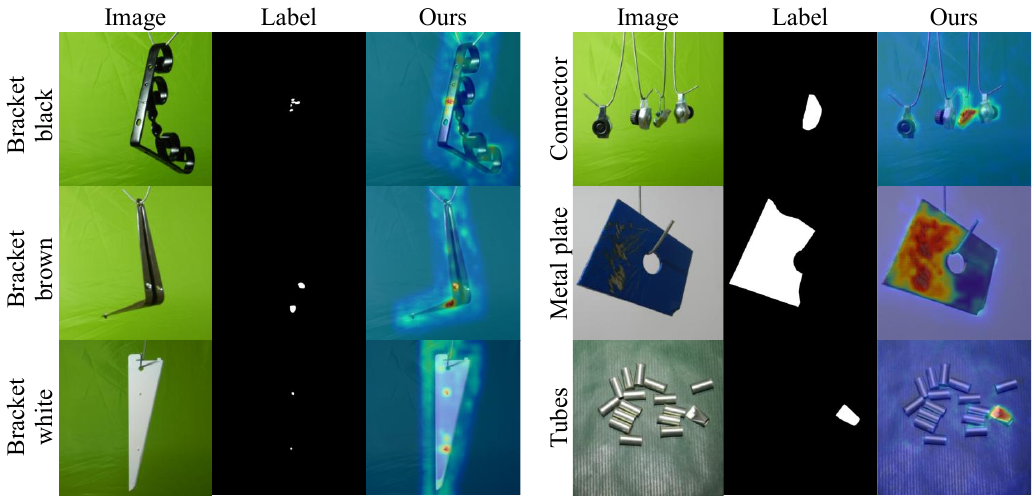} 
    \caption{Anomaly localization visualization on MPDD.}
    \label{fig:fig_mpdd_Qualitative_Comparison}
    \vspace{-4mm}
\end{figure}

\textbf{Qualitative analysis of anomaly localization:}
We conduct qualitative evaluations on MVTec-AD, VisA, and MPDD to visualize anomaly localization. As shown in Fig.~\ref{fig:fig_Qualitative_Comparison} and Fig.~\ref{fig:fig_mpdd_Qualitative_Comparison}, conventional methods often suffer from over-smoothing or false activations in background regions. Conversely, our method significantly enhances the distinction between normal and abnormal regions, generating more coherent and visually precise anomaly maps with well-defined and sharp boundaries across diverse object categories. More detailed visualizations for MVTec-AD and VisA are presented in the Supplementary.

\textbf{Qualitative analysis of matching correlation:} To illustrate the effectiveness and specific scope of cross-modal matching, we employ Gradient-weighted Class Activation Mapping (Grad-CAM) \cite{ref_gradcam} to provide a visual explanation of the bidirectional matching mechanism. Grad-CAM highlights the regions of the input image that are most influential in the matching-based decision-making process. It utilizes the gradients flowing into the normalization layer within W-MCA to generate a coarse localization map that emphasizes important regions. We take the inverse of this map to represent the degree of cross-modal matching.

\begin{figure}[t]
    \centering
    \includegraphics[width=0.48\textwidth, page=1]{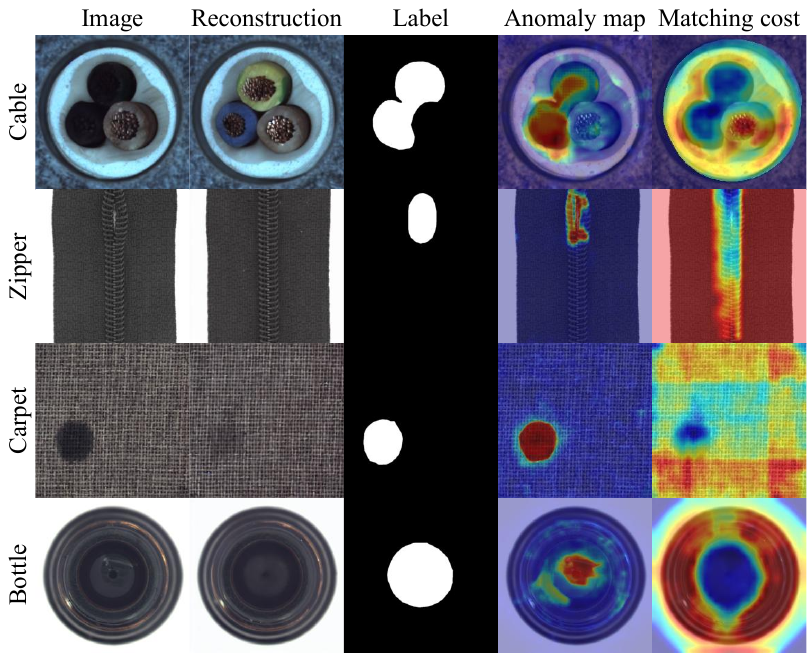} 
    \caption{Matching correlation visualization on MVTec-AD. From left to right are input images, reconstructed images, ground truth, and anomaly maps, matching correlation maps generated by our method.}
    \label{fig:attention_map}
    \vspace{-2mm}
\end{figure}

As shown in Fig.~\ref{fig:attention_map}, the last column presents the matching correlation scores, where the first two rows correspond to forward matching and the last two to backward matching. In this column, high-score regions are highlighted in red, while low-score areas are shown in blue. It is evident that normal regions yield high matching scores, whereas anomalous regions are marked by significantly lower scores. This highlights the strong discriminative capability of the attention-guided cross-modal matching in distinguishing between normal and abnormal regions.

\subsection{Ablation Studies}
To further assess the contribution of core components within our framework, we perform comprehensive ablation studies on the pseudo cross-modal setting, bidirectional matching, W-MSA, W-MCA, FCF, high-frequency information and the multi-branch decoder, using lightweight variant XMatchAD$^*$. These experiments are carried out on the MVTec-AD, VisA, and MPDD datasets using I-AUROC, P-AUROC, and P-AP, as summarized Table~\ref{tab:tab_ablation2} to Table~\ref{tab:tab_ablation_loss}. Note that the results are averaged across all categories.

\subsubsection{Pseudo cross-modal setting} In our method, we treat the input data and its reconstructed counterpart as two distinct modalities, and achieve anomaly detection through cross-modal matching and fusion. To rigorously evaluate the effectiveness of the cross-modal configuration, we construct two types of baselines.

First, we implement a conventional joint representation learning strategy like \cite{ref18} \cite{ref19} for anomaly detection, wherein the input sample and its corresponding reconstruction are processed collectively (i.e., Concat$(x_r,x_t)$). Specifically, we adopt the following settings: (i) Input fusion: the input and reconstructed images are concatenated along the channel dimension to form a unified representation; (ii) Matching adjustment: the cross-modal matching component is removed, while the W-MSA is retained to preserve local spatial dependencies; (iii) Loss function: since $\mathcal{L}_{r \rightarrow t}$ and $\mathcal{L}_{t \rightarrow r}$ are equivalent, no changes are required.
Second, we further introduce standard feature-level cosine similarity baseline (Cos. Sim.), which is derived from our method by direct computing feature-level cosine similarity between the input and reconstruction, without introducing any additional modules.

\begin{table}[t]
    \centering
    \caption{Analysis of different input formulations on MVTec-AD, VisA, and MPDD datasets, evaluated using I-AUROC, P-AUROC, and P-AP metrics.}
    \label{tab:tab_ablation2}
    \begin{tabular}{c|c|c|c}
        \toprule
        \scriptsize{Input form} & MVTec-AD & VisA & MPDD \\
        \midrule
        Concat & 96.7 / 98.3 / 49.6 & 97.3 / 97.9 / 50.9  & 96.7 / 98.9 / 58.1 \\
        Cos. Sim. & 99.5 / 98.1 / 69.0 & 98.3 / 96.9 / 47.5 & 96.3 / 98.8 / 56.0\\
        Ours & \textbf{99.6} / \textbf{99.0} / \textbf{79.2} & \textbf{98.7} / \textbf{99.3} / \textbf{55.2} & \textbf{97.5} / \textbf{99.2} / \textbf{60.2} \\
        \bottomrule
    \end{tabular}
    \vspace{-4mm}
\end{table}

\begin{table}[t]
    \centering
    \caption{Ablation study of bidirectional matching strategy on VisA using I-AUROC/P-AUROC/P-AP.}
    \label{tab:tab_ablation_bi}
    \tabcolsep=0.52cm
    \begin{tabular}{c|c|c|>{\centering\arraybackslash}p{2cm}} 
        \toprule
        ID & Forward & Backward & Results  \\
        \midrule
        1 & - & - & 97.9 / 98.3 / 48.4 \\
        2 & \checkmark & - & 97.7 / 98.8 / 52.2  \\
        3 & - & \checkmark &  98.0 / 98.9 / 52.8\\
        4 & \checkmark & \checkmark & \textbf{98.7} / \textbf{99.3} / \textbf{55.2} \\
        \bottomrule
    \end{tabular}
    \vspace{-1mm}
\end{table}



\begin{table}[t]
    \centering
    \footnotesize
    \caption{Ablation studies of the modules in our method on the VisA dataset using I-AUROC/P-AUROC.}
    \label{tab:tab_ablation1}
    \tabcolsep=0.1cm
    \begin{tabular}{ccccccc
    } 
        \toprule
         \multirow{2}{*}{ID} & \multirow{2}{*}{W-MSA} & \multirow{2}{*}{W-MCA} & \multirow{2}{*}{FCF} & \multicolumn{3}{c}{Performances}  \\
        \cmidrule{5-7}
        &&&&FLOPs (G) & Params. (M) & Results\\
        \midrule
         1&- & -&-&104.82 {\scriptsize(+0)}& 148.01 {\scriptsize(+0)} & 98.3 / 96.9  \\
         2&- & $\checkmark$ & $\checkmark$ & 119.16 {\scriptsize(+14.34)} & 162.29 {\scriptsize(+14.28)} & 98.4 / 98.7 \\
         3&$\checkmark$ & -& $\checkmark$ & 119.16 {\scriptsize(+14.34)} & 162.29 {\scriptsize(+14.28)} & 97.9 / 98.3 \\
         4&$\checkmark$ & $\checkmark$ &- & 119.43 {\scriptsize(+14.61)} & 164.54 {\scriptsize(+16.53)} & 98.5 / 98.7\\
         5&$\checkmark$ & -& -& 119.16 {\scriptsize(+14.34)} & 162.15 {\scriptsize(+14.14)} & 98.1  / 98.1 \\
         6&- & $\checkmark$ & -&  119.16 {\scriptsize(+14.34)} & 162.15 {\scriptsize(+14.14)} & 98.4 / 98.7 \\
         7&$\checkmark$ & $\checkmark$ & $\checkmark$ & 119.43 {\scriptsize(+14.61)} & 164.68 {\scriptsize(+16.67)} & \textbf{98.7} / \textbf{99.3}\\
        \bottomrule
    \end{tabular}
    \begin{tablenotes}
\footnotesize
\item[1] FCF: Frequency-aware Cross-modal Fusion.
\end{tablenotes}
    \vspace{-1mm}
\end{table}


\begin{table}[t]
    \centering
    \footnotesize
    \caption{Ablation study of cosine similarity and W-MCA on VisA using I-AUROC/I-AP/P-AUROC.}
    \label{tab:tab_cos_matching}
    \tabcolsep=0.5cm
    \begin{tabular}{c|l|c
    } 
        \toprule
        ID &  Settings &  Results \\
        \midrule
         1&only W-MCA & 98.0 / 98.6 / 97.3  \\
         2&only cosine similarity & 98.3 / 98.5 / 96.9\\
         \midrule
         3&W-MCA $\rightarrow$ cosine similarity & 97.9 / 98.3 / 98.5\\
         4& W-MCA $\rightarrow$ G-MCA & 97.7 / 98.4 / 98.8\\
         5&Ours & \textbf{98.7} / \textbf{99.0} / \textbf{99.3}\\
        \bottomrule
    \end{tabular}
    \begin{tablenotes}
\footnotesize
\item[1] G-MCA: Global Multi-head Cross Attention.
\end{tablenotes}
    \vspace{-1mm}
\end{table}



\begin{table}[t]
    \centering
    \caption{Ablation study on the impact of different frequency components in the FCF on VisA using I-AUROC/P-AUROC/P-AP.}
    \label{tab:tab_ablation_LL}
    \tabcolsep=0.42cm
    \begin{tabular}{l|c c>{\centering\arraybackslash}p{1cm}} 
        \toprule
         Settings & I-AUROC & P-AUROC & P-AP  \\
        \midrule
         only Low freq. & 98.4 & 98.6 &  52.6 \\
        w/ Low freq. &98.3 & 99.0 & 54.0  \\
        w/o Low freq. & \textbf{98.7} & \textbf{99.3} & \textbf{55.2} \\
        \bottomrule
    \end{tabular}
    \vspace{-1mm}
\end{table}

As presented in Table~\ref{tab:tab_ablation2}, the pseudo cross-modal model significantly outperforms all variants across multiple evaluation metrics on the MVTec-AD, VisA, and MPDD datasets. Notably, the variant with simple concatenation exhibits a noticeable performance drop, indicating the importance of explicit cross-modal matching in capturing inter-modal relationships. These results confirm that the proposed cross-modal formulation contributes substantially to the overall effectiveness of the framework.

\subsubsection{Effectiveness of Bidirectional matching}
We conduct ablation experiments on VisA to investigate the effectiveness of the bidirectional matching introduced in Section \ref{sec:matching}. As shown in Table \ref{tab:tab_ablation_bi}, four variants are evaluated: ID1 (without matching), ID2 (backward matching only), ID3 (forward matching only), and ID4 (bidirectional matching). It can be observed that using only forward or backward matching results in lower performance compared to bidirectional matching. This suggests that the bidirectional matching mechanism works more effectively in a collaborative manner, jointly enhancing the model's ability to identify anomalies.

\subsubsection{Effectiveness of window-based multi-head self-attention}
To assess the effectiveness of the W-MSA component introduced in Section \ref{sec:MSA}, we remove it from the attention-guided cross-modal matching module while retaining only the W-MCA. As shown in Table \ref{tab:tab_ablation1} (ID2 vs. ID7), the performance of our method with W-MSA (ID2) is notably superior to that without W-MSA (ID7), while incurring only a marginal increase in model complexity (e.g., 0.3 GFLOPs and 2.4M parameters), which further highlights the critical role of W-MSA in enhancing intra-modal feature representations.

\subsubsection{Effectiveness of window-based multi-head cross-attention} We conduct ablation studies on VisA to evaluate the effectiveness of the W-MCA from three perspectives: (i) mechanism comparison, (ii) component replacement, and (iii) module removal. Specifically, we compare W-MCA with its global counterpart (G-MCA), cosine-similarity-based matching, and the setting without the matching module.

(i) \textbf{Mechanism comparison.} We compare pure feature-level cosine similarity (only cosine similarity) with our attention-guided matching (only W-MCA) in Table \ref{tab:tab_cos_matching} (ID1 vs. ID2). The cosine-only variant achieves comparable I-AUROC (98.3\% vs. 98.0\%) but inferior P-AUROC/I-AP (96.9\%/98.5\% vs. 97.3\%/98.6\%), indicating that attention-guided matching enables more effective feature alignment than direct similarity computation.

(ii) \textbf{Component replacement.} (a) We first replace W-MCA with cosine similarity (W-MCA $\rightarrow$ cosine similarity) in Table \ref{tab:tab_cos_matching}. This substitution leads to a noticeable performance drop of 0.8\%/0.7\%/0.8\% (ID3 vs. ID5) in I-AUROC/I-AP/P-AUROC compared with the full model, demonstrating that simple similarity metrics are insufficient for accurate cross-modal matching. (b) We further replace W-MCA with G-MCA (W-MCA $\rightarrow$ G-MCA) in Table \ref{tab:tab_cos_matching} (ID4 vs. ID5) to assess the effectiveness of local versus global matching mechanisms. G-MCA performs cross-modal attention over the entire feature map without spatial constraints or window partitioning. In contrast, W-MCA restricts attention to local windows, emphasizing spatially localized interactions. The results show that the global variant leads to performance drops of 1.0\%/0.6\%/0.5\% in I-AUROC/I-AP/P-AUROC. This observation suggests that local matching is more effective for capturing fine-grained structures critical for distinguishing subtle anomalies, whereas global attention tends to dilute such local discrepancies due to increased feature interactions.

(iii) \textbf{Module removal.} We remove W-MCA entirely to further assess its contribution. As shown in Table~\ref{tab:tab_ablation1} (ID3 vs. ID7), incorporating W-MCA improves performance by 0.8\%/1.0\% in I-AUROC/P-AUROC, with only marginal computational overhead, highlighting its essential role in the proposed XMatchAD framework.

Overall, these results consistently demonstrate that W-MCA provides more effective cross-modal matching and that its performance gains are neither incidental nor easily replicated by alternative components.

\begin{table}[t]
    \centering
    \footnotesize
    \caption{Impact of loss terms (multi-branch decoder) on anomaly detection performance and model complexity over MVTec-AD using I-AUROC/P-AUROC/P-AP.}
    \label{tab:tab_ablation_loss}
    \tabcolsep=0.1cm
    \begin{tabular}{cccccccc
    } 
        \toprule
        \multirow{2}{*}{ID} & \multirow{2}{*}{$\mathcal{L}_f$} &\multirow{2}{*}{$\mathcal{L}_{t  \to r}$} & \multirow{2}{*}{$\mathcal{L}_{r \to t}$} &  \multicolumn{3}{c}{Performances} \\
        \cmidrule{5-7}
         &&&&FLOPs (G) & Params. (M) & Results \\
        \midrule 
         1&- & -&-&104.82 {\scriptsize(+0)}& 148.01 {\scriptsize(+0)} & 99.8 / 98.1 / 69.0\\
         2& - & $\checkmark$ & - & 118.53 \scriptsize{(+13.71)} & 160.23 \scriptsize{(+12.22)} & 99.2 / 98.3 / 71.1\\
         3&- & - & $\checkmark$ &118.53 \scriptsize{(+13.71)} & 160.23 \scriptsize{(+12.22)} & 98.8 / 98.0 / 71.8\\
         4& $\checkmark$ & - & - & 119.06 \scriptsize{(+14.24)} & 163.82 \scriptsize{(+15.81)}& 98.9	/ 98.3 / 71.2\\
         5& - & $\checkmark$ & $\checkmark$ &  118.72 \scriptsize{(+13.90)} & 162.82 \scriptsize{(+14.81)} & 99.4 / 98.6 / 74.9\\
         6& $\checkmark$ & - & $\checkmark$ & 119.24 \scriptsize{(+14.42)} & 164.25 \scriptsize{(+16.24)} & 99.2 / 98.4 / 72.3\\
         7&$\checkmark$ & $\checkmark$ & - & 119.24 \scriptsize{(+14.42)} & 164.25 \scriptsize{(+16.24)} & 99.6 / 96.9 / 73.8\\
         8&$\checkmark$ & $\checkmark$ & $\checkmark$ & 119.43 \scriptsize{(+14.61)} & 164.68 \scriptsize{(+16.67)} & \textbf{99.8} / \textbf{99.0} / \textbf{79.2}\\
        \bottomrule
    \end{tabular}
    \vspace{-1mm}
\end{table}


\subsubsection{Frequency-aware cross-modal fusion} To evaluate the effectiveness of FCF, we compare configurations with and without FCF under similar settings on VisA as shown in Table \ref{tab:tab_ablation1}. Specifically, comparing ID5 and ID3, the introduction of FCF improves the performance from 98.1\% to 98.3\% in P-AUROC, demonstrating its ability to enhance pixel-level localization. Furthermore, when all components are combined (ID 7), incorporating FCF contributes to achieving the best overall performance (98.7\%/99.3\%) while introducing only a negligible increase in model parameters (ID4 vs. ID7). These results indicate that FCF provides complementary benefits and effectively enhances feature fusion for anomaly detection. These results validate the effectiveness of the proposed FCF component and demonstrate its contribution to more accurate anomaly localization.

\subsubsection{Effectiveness of high-frequency components in FCF}
To verify the contribution of high-frequency components (Eqn.~(\ref{eq:hl})) for cross-modal fusion, we design three cases: the original XMatchAD with only low-frequency components (only Low freq.), with low-frequency components added (w/ Low freq.) and without low-frequency components (w/o Low freq.). As presented in Table \ref{tab:tab_ablation_LL}, the model performs worst when relying solely on low-frequency information. Although incorporating low-frequency components (w/ Low freq.) yields improved results, the performance remains inferior to that of the configuration with only high-frequency information. These findings underscore the greater significance of high-frequency components, which can be attributed to the idea that spatial detail is contained in high frequencies.


\subsubsection{Effectiveness of the multi-branch decoder} We conduct a detailed ablation on the decoder design ($\mathcal{L}_{t \to r}$, $\mathcal{L}_{r \to t}$ and $\mathcal{L}_{f}$) in Table \ref{tab:tab_ablation_loss}. Using any single branch alone (ID2-ID4) yields noticeable improvements of 2.1\%, 2.8\%, and 2.2\% in P-AP compared to the baseline (ID1), which is derived from our
method by direct computing feature-level cosine similarity
between the input and reconstruction, while introducing only limited parameter overhead (12.22M, 12.22M, and 15.81M, respectively). Combining branches progressively (ID5-ID7) consistently improves performance, indicating that different branches capture complementary information. And the full multi-branch design (ID8) achieves the best performance (99.8\%/99.0\%/79.2\% in I-AP/P-AUROC/P-AP), while the increase in parameters and FLOPs remains moderate (+16.67M/+14.61G). These results consistently suggest that the collaborative learning of latent representations through multiple loss constraints has a positive effect on model performance, while the multi-branch design introduces only modest computational overhead.

\begin{table}[t]
\centering
    \footnotesize
    \caption{Computational efficiency of our method, including FLOPs, memory usage, number of parameters, and inference time.}
    \label{tab:tab_computation_complex}
    \tabcolsep=0.12cm
    \begin{tabular}{lccccc
    } 
        \toprule
        Methods & FLOPs & Mem. (GB) & Params. (M) & Inf. (s) \\
        \midrule
        XMatchAD & $>$2.2T& 11.65 & 16.58 & 4.190\\
        XMatchAD$^*$ & 119.43G & 1.19 & 164.68 &  0.184  \\
        \bottomrule
    \end{tabular}
    \vspace{-1mm}
\end{table}

\subsection{Computational Efficiency}\label{sec:efficient}
We report the FLOPs, memory usage, number of parameters, and inference time for our method in Table \ref{tab:tab_computation_complex}. Specifically, XMatchAD$^*$ requires 119.43G FLOPs with an inference time of 0.184s per image, whereas XMatchAD incurs over 2.2T FLOPs and 4.19s per image. For comparison, the reconstruction methods Dinomaly requires 104.82G FLOPs with an inference time of 0.04s per image, while GLAD exceeds 2.2T FLOPs with 3.96s per image. These results suggest that the computational overhead of our method primarily stems from the reconstruction process.

\section{Conclusion}
In this paper, we propose XMatchAD, a novel framework for multi-class UAD, which reformulates the task as a pseudo cross-modal matching problem. Specifically, the input image and its reconstructed counterpart are regarded as two complementary modalities, providing precise identification of anomalies with various shapes and structures by matching inter-modal dependency relationships across multiple feature levels. To this end, we introduce a local attention-guided cross-modal matching mechanism that aligns representations between modalities, refines feature quality, and effectively captures fine-grained anomalies. Moreover, an adaptive frequency-aware fusion module is integrated to further improve the accuracy of boundary delineation. Extensive experiments on three benchmarks verify that our method achieves SOTA performance, underscoring its strong effectiveness and robustness in multi-class UAD.

\textbf{Limitations and future work.}
A limitation of our method is that the inference time depends directly on the speed of reconstruction, with faster reconstruction leading to shorter inference time. Therefore, in future work, we will focus more on embedding-based cross-modal anomaly detection to mitigate the negative impact of reconstruction and further boost the sensitivity to inconspicuous or low-contrast anomalies.

\bibliographystyle{IEEEtran}
\bibliography{cas-refs}

\begin{thebibliography}{10}
\providecommand{\url}[1]{#1}
\csname url@samestyle\endcsname
\providecommand{\newblock}{\relax}
\providecommand{\bibinfo}[2]{#2}
\providecommand{\BIBentrySTDinterwordspacing}{\spaceskip=0pt\relax}
\providecommand{\BIBentryALTinterwordstretchfactor}{4}
\providecommand{\BIBentryALTinterwordspacing}{\spaceskip=\fontdimen2\font plus
\BIBentryALTinterwordstretchfactor\fontdimen3\font minus
  \fontdimen4\font\relax}
\providecommand{\BIBforeignlanguage}[2]{{%
\expandafter\ifx\csname l@#1\endcsname\relax
\typeout{** WARNING: IEEEtran.bst: No hyphenation pattern has been}%
\typeout{** loaded for the language `#1'. Using the pattern for}%
\typeout{** the default language instead.}%
\else
\language=\csname l@#1\endcsname
\fi
#2}}
\providecommand{\BIBdecl}{\relax}
\BIBdecl

\bibitem{ref1}
D.~Tri~Phan, V.~Hoang Minh~Doan, J.~Choi, B.~Lee, and J.~Oh, ``Aadc-net: A
  multimodal deep learning framework for automatic anomaly detection in
  real-time surveillance,'' \emph{IEEE Transactions on Instrumentation and
  Measurement}, vol.~74, pp. 1--13, 2025.

\bibitem{ref2}
Y.~Liang, J.~Zhang, S.~Zhao, R.~Wu, Y.~Liu, and S.~Pan, ``Omni-frequency
  channel-selection representations for unsupervised anomaly detection,''
  \emph{IEEE Transactions on Image Processing}, vol.~32, pp. 4327--4340, 2023.

\bibitem{ref3}
S.~Lu, W.~Zhang, H.~Zhao, H.~Liu, N.~Wang, and H.~Li, ``Anomaly detection for
  medical images using heterogeneous auto-encoder,'' \emph{IEEE Transactions on
  Image Processing}, vol.~33, pp. 2770--2782, 2024.

\bibitem{ref4}
Z.~Huang, B.~Zhang, G.~Hu, L.~Li, Y.~Xu, and Y.~Jin, ``Enhancing unsupervised
  anomaly detection with score-guided network,'' \emph{IEEE Transactions on
  Neural Networks and Learning Systems}, vol.~35, no.~10, pp. 14\,754--14\,769,
  2023.

\bibitem{ref5}
T.~Xiang, Y.~Zhang, Y.~Lu, A.~Yuille, C.~Zhang, W.~Cai, and Z.~Zhou,
  ``Exploiting structural consistency of chest anatomy for unsupervised anomaly
  detection in radiography images,'' \emph{IEEE Transactions on Pattern
  Analysis and Machine Intelligence}, vol.~46, no.~9, pp. 6070--6081, 2024.

\bibitem{ref6}
H.~Liu and J.~Sun, ``Unistad: An unified triple-tower student–teacher model
  for multi-class anomaly detection and localization,'' \emph{IEEE Transactions
  on Circuits and Systems for Video Technology}, vol.~35, no.~4, pp.
  3196--3208, 2025.

\bibitem{ref7}
Y.~Cheng, Y.~Cao, D.~Wang, W.~Shen, and W.~Li, ``Boosting global-local feature
  matching via anomaly synthesis for multi-class point cloud anomaly
  detection,'' \emph{IEEE Transactions on Automation Science and Engineering},
  vol.~22, pp. 12\,560--12\,571, 2025.

\bibitem{ref8}
Z.~Zhang, M.~Cai, H.~Wang, G.~Wu, T.~Chai, and X.~Zhu, ``Costfilter-ad:
  Enhancing anomaly detection through matching cost filtering,'' \emph{arXiv
  preprint arXiv:2505.01476}, 2025.

\bibitem{ref9}
T.~D. Tien, A.~T. Nguyen, N.~H. Tran, T.~D. Huy, S.~Duong, C.~D.~T. Nguyen, and
  S.~Q. Truong, ``Revisiting reverse distillation for anomaly detection,'' in
  \emph{CVPR}, 2023, pp. 24\,511--24\,520.

\bibitem{ref10}
Q.~Chen, H.~Luo, C.~Lv, and Z.~Zhang, ``A unified anomaly synthesis strategy
  with gradient ascent for industrial anomaly detection and localization,'' in
  \emph{ECCV}, 2024, pp. 37--54.

\bibitem{ref11}
Y.~Shi, J.~Yang, and Z.~Qi, ``Unsupervised anomaly segmentation via deep
  feature reconstruction,'' \emph{Neurocomputing}, vol. 424, pp. 9--22, 2021.

\bibitem{ref12}
W.~Liu, R.~Li, M.~Zheng, S.~Karanam, Z.~Wu, B.~Bhanu, R.~J. Radke, and
  O.~Camps, ``Towards visually explaining variational autoencoders,'' in
  \emph{CVPR}, 2020, pp. 8642--8651.

\bibitem{ref13}
J.~Hou, Y.~Zhang, Q.~Zhong, D.~Xie, S.~Pu, and H.~Zhou, ``Divide-and-assemble:
  Learning block-wise memory for unsupervised anomaly detection,'' in
  \emph{ICCV}, 2021, pp. 8791--8800.

\bibitem{ref_shortcut_input}
D.~Gong, L.~Liu, V.~Le, B.~Saha, M.~R. Mansour, S.~Venkatesh, and A.~v.~d.
  Hengel, ``Memorizing normality to detect anomaly: Memory-augmented deep
  autoencoder for unsupervised anomaly detection,'' in \emph{ICCV}, 2019, pp.
  1705--1714.

\bibitem{ref14}
Y.~Liang, J.~Zhang, S.~Zhao, R.~Wu, Y.~Liu, and S.~Pan, ``Omni-frequency
  channel-selection representations for unsupervised anomaly detection,''
  \emph{IEEE Transactions on Image Processing}, vol.~32, pp. 4327--4340, 2023.

\bibitem{ref15}
V.~Zavrtanik, M.~Kristan, and D.~Sko{\v{c}}aj, ``Reconstruction by inpainting
  for visual anomaly detection,'' \emph{Pattern Recognition}, vol. 112, p.
  107706, 2021.

\bibitem{ref_25}
X.~Zhang, N.~Li, J.~Li, T.~Dai, Y.~Jiang, and S.-T. Xia, ``Unsupervised surface
  anomaly detection with diffusion probabilistic model,'' in \emph{ICCV}, 2023,
  pp. 6782--6791.

\bibitem{ref30}
H.~Yao, M.~Liu, Z.~Yin, Z.~Yan, X.~Hong, and W.~Zuo, ``Glad: Towards better
  reconstruction with global and local adaptive diffusion models for
  unsupervised anomaly detection,'' in \emph{ECCV}, 2024, pp. 1--17.

\bibitem{ref31}
H.~He, J.~Zhang, H.~Chen, X.~Chen, Z.~Li, X.~Chen, Y.~Wang, C.~Wang, and
  L.~Xie, ``A diffusion-based framework for multi-class anomaly detection,'' in
  \emph{AAAI}, vol.~38, no.~8, 2024, pp. 8472--8480.

\bibitem{ref_shortcut}
Z.~You, K.~Yang, W.~Luo, L.~Cui, Y.~Zheng, and X.~Le, ``Adtr: Anomaly detection
  transformer with feature reconstruction,'' in \emph{ICONIP}, 2022, pp.
  298--310.

\bibitem{ref18}
V.~Zavrtanik, M.~Kristan, and D.~Sko{\v{c}}aj, ``Draem-a discriminatively
  trained reconstruction embedding for surface anomaly detection,'' in
  \emph{ICCV}, 2021, pp. 8330--8339.

\bibitem{ref19}
H.~Zhang, Z.~Wang, D.~Zeng, Z.~Wu, and Y.-G. Jiang, ``Diffusionad: Norm-guided
  one-step denoising diffusion for anomaly detection,'' \emph{IEEE Transactions
  on Pattern Analysis and Machine Intelligence}, vol.~47, no.~8, pp.
  7140--7152, 2025.

\bibitem{ref21}
V.~Zavrtanik, M.~Kristan, and D.~Sko{\v{c}}aj, ``Dsr--a dual subspace
  re-projection network for surface anomaly detection,'' in \emph{ECCV}, 2022,
  pp. 539--554.

\bibitem{ref25-mdps}
D.~Wu, S.~Fan, X.~Zhou, L.~Yu, Y.~Deng, J.~Zou, and B.~Lin, ``Unsupervised
  anomaly detection via masked diffusion posterior sampling,'' \emph{arXiv
  preprint arXiv:2404.17900}, 2024.

\bibitem{ref25-ddad}
A.~Mousakhan, T.~Brox, and J.~Tayyub, ``Anomaly detection with conditioned
  denoising diffusion models,'' in \emph{DAGM German Conference on Pattern
  Recognition}, 2024, pp. 181--195.

\bibitem{ref25-uniad}
Z.~You, L.~Cui, Y.~Shen, K.~Yang, X.~Lu, Y.~Zheng, and X.~Le, ``A unified model
  for multi-class anomaly detection,'' \emph{NeurIPS}, vol.~35, pp. 4571--4584,
  2022.

\bibitem{ref23}
X.~Zhang, S.~Li, X.~Li, P.~Huang, J.~Shan, and T.~Chen, ``Destseg: Segmentation
  guided denoising student-teacher for anomaly detection,'' in \emph{CVPR},
  2023, pp. 3914--3923.

\bibitem{ref24}
Y.~Zhao, ``Just noticeable learning for unsupervised anomaly localization and
  detection,'' in \emph{ICME}, 2022, pp. 01--06.

\bibitem{ref27}
K.~Amolins, Y.~Zhang, and P.~Dare, ``Wavelet based image fusion techniques—an
  introduction, review and comparison,'' \emph{ISPRS Journal of photogrammetry
  and Remote Sensing}, vol.~62, no.~4, pp. 249--263, 2007.

\bibitem{ref29}
X.~Yan, H.~Zhang, X.~Xu, X.~Hu, and P.-A. Heng, ``Learning semantic context
  from normal samples for unsupervised anomaly detection,'' in \emph{AAAI},
  vol.~35, no.~4, 2021, pp. 3110--3118.

\bibitem{ref32}
C.~Cui, Y.~Ma, X.~Cao, W.~Ye, Y.~Zhou, K.~Liang, J.~Chen, J.~Lu, Z.~Yang, K.-D.
  Liao \emph{et~al.}, ``A survey on multimodal large language models for
  autonomous driving,'' in \emph{WACV}, 2024, pp. 958--979.

\bibitem{ref33}
K.~Dasgupta, A.~Das, S.~Das, U.~Bhattacharya, and S.~Yogamani,
  ``Spatio-contextual deep network-based multimodal pedestrian detection for
  autonomous driving,'' \emph{IEEE transactions on intelligent transportation
  systems}, vol.~23, no.~9, pp. 15\,940--15\,950, 2022.

\bibitem{ref34}
Z.~Zhang, G.~Wu, J.~Zhang, X.~Zhu, D.~Tao, and T.~Chai, ``Unified domain
  adaptive semantic segmentation,'' \emph{IEEE Transactions on Pattern Analysis
  and Machine Intelligence}, vol.~47, no.~8, pp. 6731--6748, 2025.

\bibitem{ref35}
T.~Yu, K.~Fu, J.~Zhang, Q.~Huang, and J.~Yu, ``Multi-granularity contrastive
  cross-modal collaborative generation for end-to-end long-term video question
  answering,'' \emph{IEEE Transactions on Image Processing}, vol.~33, pp.
  3115--3129, 2024.

\bibitem{ref36}
S.~Chen, P.-L. Guhur, C.~Schmid, and I.~Laptev, ``History aware multimodal
  transformer for vision-and-language navigation,'' \emph{NeurIPS}, vol.~34,
  pp. 5834--5847, 2021.

\bibitem{ref37}
G.~Georgakis, K.~Schmeckpeper, K.~Wanchoo, S.~Dan, E.~Miltsakaki, D.~Roth, and
  K.~Daniilidis, ``Cross-modal map learning for vision and language
  navigation,'' in \emph{CVPR}, 2022, pp. 15\,460--15\,470.

\bibitem{ref38}
F.~Behrad and M.~S. Abadeh, ``An overview of deep learning methods for
  multimodal medical data mining,'' \emph{Expert Systems with Applications},
  vol. 200, p. 117006, 2022.

\bibitem{ref39}
H.~Diao, Y.~Zhang, W.~Liu, X.~Ruan, and H.~Lu, ``Plug-and-play regulators for
  image-text matching,'' \emph{IEEE Transactions on Image Processing}, vol.~32,
  pp. 2322--2334, 2023.

\bibitem{ref40}
Y.~Ding, X.~Yu, and Y.~Yang, ``Rfnet: Region-aware fusion network for
  incomplete multi-modal brain tumor segmentation,'' in \emph{ICCV}, 2021, pp.
  3975--3984.

\bibitem{ref41}
Z.~Zhao, H.~Bai, J.~Zhang, Y.~Zhang, K.~Zhang, S.~Xu, D.~Chen, R.~Timofte, and
  L.~Van~Gool, ``Equivariant multi-modality image fusion,'' in \emph{CVPR},
  2024, pp. 25\,912--25\,921.

\bibitem{ref_unet}
O.~Ronneberger, P.~Fischer, and T.~Brox, ``U-net: Convolutional networks for
  biomedical image segmentation,'' in \emph{MICCAI}, 2015, pp. 234--241.

\bibitem{ref42}
S.~W. Zamir, A.~Arora, S.~Khan, M.~Hayat, F.~S. Khan, and M.-H. Yang,
  ``Restormer: Efficient transformer for high-resolution image restoration,''
  in \emph{CVPR}, 2022, pp. 5728--5739.

\bibitem{ref43}
X.~Li, B.~Fan, J.~Tian, and H.~Fan, ``Gafusion: Adaptive fusing lidar and
  camera with multiple guidance for 3d object detection,'' in \emph{CVPR},
  2024, pp. 21\,209--21\,218.

\bibitem{ref44}
H.~Li, J.~Wang, J.~Yuan, Y.~Li, W.~Weng, Y.~Peng, Y.~Zhang, Z.~Xiong, and
  X.~Sun, ``Event-assisted low-light video object segmentation,'' in
  \emph{CVPR}, 2024, pp. 3250--3259.

\bibitem{ref45}
D.~Jiang and M.~Ye, ``Cross-modal implicit relation reasoning and aligning for
  text-to-image person retrieval,'' in \emph{CVPR}, 2023, pp. 2787--2797.

\bibitem{ref46}
Y.~Lu, Q.~Jiang, R.~Chen, Y.~Hou, X.~Zhu, and Y.~Ma, ``See more and know more:
  Zero-shot point cloud segmentation via multi-modal visual data,'' in
  \emph{ICCV}, 2023, pp. 21\,674--21\,684.

\bibitem{ref47}
J.~Li, R.~Selvaraju, A.~Gotmare, S.~Joty, C.~Xiong, and S.~C.~H. Hoi, ``Align
  before fuse: Vision and language representation learning with momentum
  distillation,'' \emph{NeurIPS}, vol.~34, pp. 9694--9705, 2021.

\bibitem{ref48}
M.~Caron, H.~Touvron, I.~Misra, H.~J{\'e}gou, J.~Mairal, P.~Bojanowski, and
  A.~Joulin, ``Emerging properties in self-supervised vision transformers,'' in
  \emph{ICCV}, 2021, pp. 9650--9660.

\bibitem{ref26}
Z.~Liu, Y.~Lin, Y.~Cao, H.~Hu, Y.~Wei, Z.~Zhang, S.~Lin, and B.~Guo, ``Swin
  transformer: Hierarchical vision transformer using shifted windows,'' in
  \emph{ICCV}, 2021, pp. 10\,012--10\,022.

\bibitem{ref53}
A.~Vaswani, N.~Shazeer, N.~Parmar, J.~Uszkoreit, L.~Jones, A.~N. Gomez,
  {\L}.~Kaiser, and I.~Polosukhin, ``Attention is all you need,''
  \emph{NeurIPS}, vol.~30, 2017.

\bibitem{ref_freq}
K.~Amolins, Y.~Zhang, and P.~Dare, ``Wavelet based image fusion techniques—an
  introduction, review and comparison,'' \emph{ISPRS Journal of photogrammetry
  and Remote Sensing}, vol.~62, no.~4, pp. 249--263, 2007.

\bibitem{ref28}
Y.~Zhou, J.~Huang, C.~Wang, L.~Song, and G.~Yang, ``Xnet: Wavelet-based low and
  high frequency fusion networks for fully-and semi-supervised semantic
  segmentation of biomedical images,'' in \emph{ICCV}, 2023, pp.
  21\,085--21\,096.

\bibitem{ref49}
Z.~Liu, Y.~Zhou, Y.~Xu, and Z.~Wang, ``Simplenet: A simple network for image
  anomaly detection and localization,'' in \emph{CVPR}, 2023, pp.
  20\,402--20\,411.

\bibitem{ref50}
J.~Zhang, X.~Chen, Y.~Wang, C.~Wang, Y.~Liu, X.~Li, M.-H. Yang, and D.~Tao,
  ``Exploring plain vit features for multi-class unsupervised visual anomaly
  detection,'' \emph{Computer Vision and Image Understanding}, vol. 253, p.
  104308, 2025.

\bibitem{ref51}
S.~Damm, M.~Laszkiewicz, J.~Lederer, and A.~Fischer, ``Anomalydino: Boosting
  patch-based few-shot anomaly detection with dinov2,'' in \emph{WACV}.\hskip
  1em plus 0.5em minus 0.4em\relax IEEE, 2025, pp. 1319--1329.

\bibitem{dinomaly}
J.~Guo, S.~Lu, W.~Zhang, F.~Chen, H.~Li, and H.~Liao, ``Dinomaly: The less is
  more philosophy in multi-class unsupervised anomaly detection,'' in
  \emph{Proceedings of the Computer Vision and Pattern Recognition Conference},
  2025, pp. 20\,405--20\,415.

\bibitem{ref52}
E.~Parzen, ``On estimation of a probability density function and mode,''
  \emph{The annals of mathematical statistics}, vol.~33, no.~3, pp. 1065--1076,
  1962.

\bibitem{ref_gradcam}
R.~R. Selvaraju, M.~Cogswell, A.~Das, R.~Vedantam, D.~Parikh, and D.~Batra,
  ``Grad-cam: Visual explanations from deep networks via gradient-based
  localization,'' in \emph{ICCV}, 2017, pp. 618--626.

\end{thebibliography}

\begin{IEEEbiography}[{\includegraphics[width=1in,height=1.25in,clip,keepaspectratio]{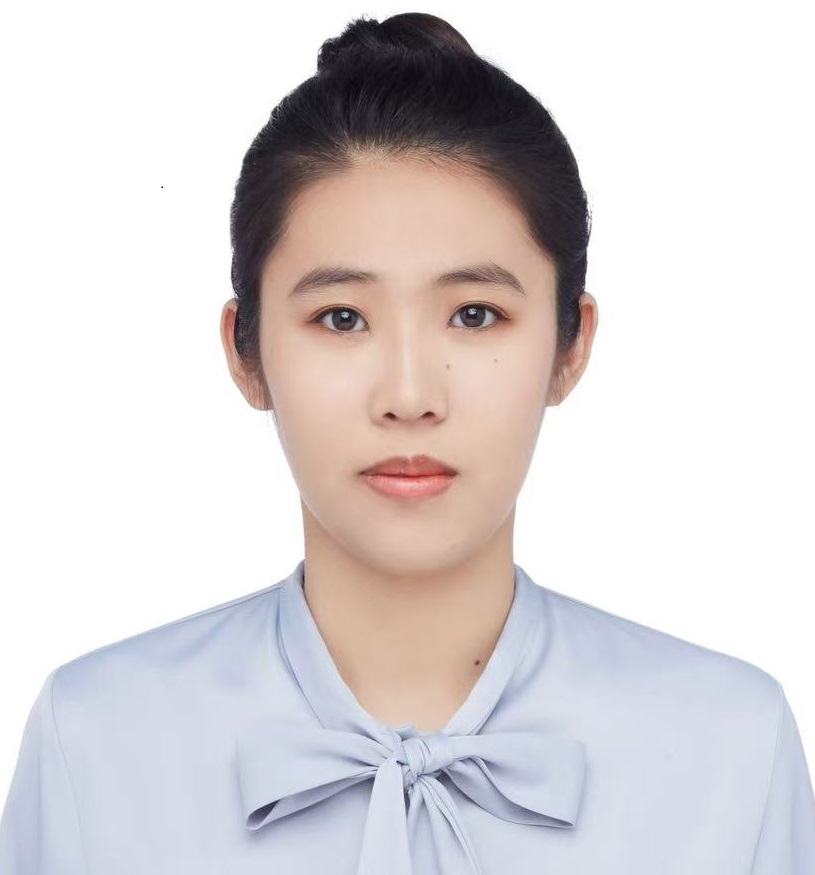}}]{Mingxiu Cai} received the MS degree from the School of Computer Science, Nanjing Audit University, China, in 2024.
	 She is currently working toward a Ph.D. degree in the State Key Laboratory of Synthetical Automation for Process Industries, Northeastern University, Shenyang, China. Her current research interests include computer vision, deep learning, multi-modal learning, and their applications in industrial fields.
\end{IEEEbiography}

\begin{IEEEbiography}[{\includegraphics[width=1in,height=1.25in,clip,keepaspectratio]{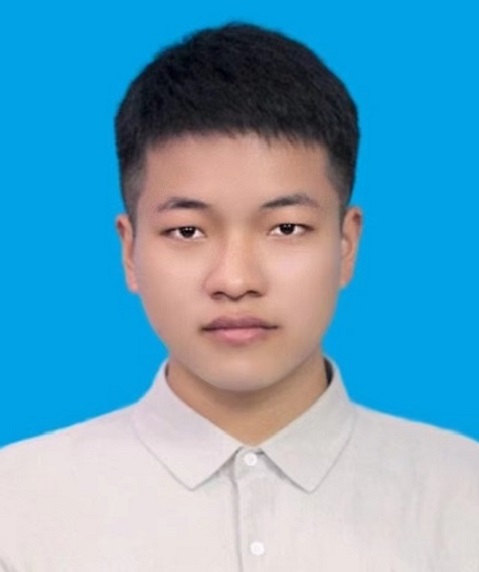}}]{Zhe Zhang}
	received the BS degree in the College of Information Science and Engineering, Northeastern University, China, in 2021. He is currently working toward a Ph.D. degree in the State Key Laboratory of Synthetical Automation for Process Industries, Northeastern University, Shenyang, China. His current research interests include computer vision, deep learning, video representation learning, multi-modal learning, and their applications in industrial fields.
\end{IEEEbiography}

\begin{IEEEbiography}[{\includegraphics[width=1in]{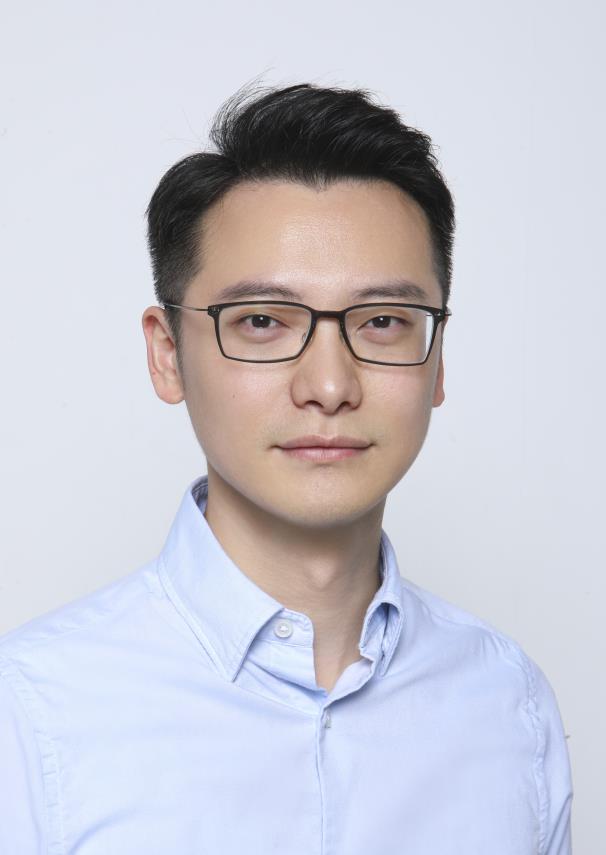}}]{Gaochang Wu} (IEEE Member)
received the BE and MS degrees in mechanical engineering in Northeastern University, Shenyang, China, in 2013 and 2015, respectively, and Ph.D. degree in control theory and control engineering in Northeastern University, Shenyang, China in 2020. He is currently an associate professor in the State Key Laboratory of Synthetical Automation for Process Industries, Northeastern University. He was selected for the 2022-2024 Youth Talent Support Program of the Chinese Association of Automation. His current research interests include multimodal perception and recognition, light field imaging and processing, and computer vision in industrial scenarios.
\end{IEEEbiography}
\vspace{-4mm}

\begin{IEEEbiography}[{\includegraphics[width=1in]{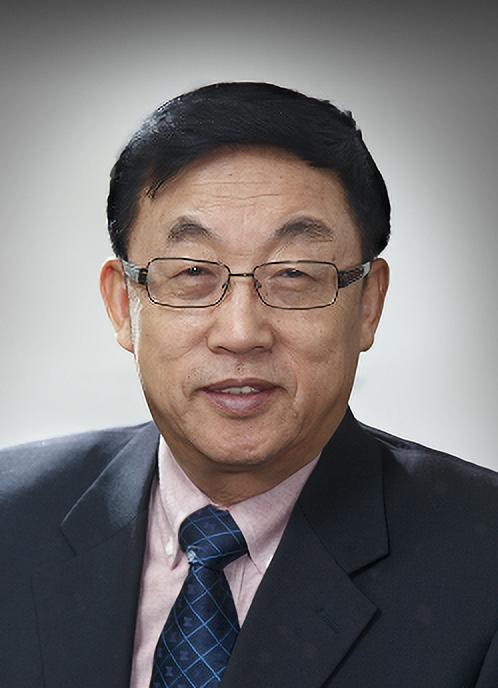}}]{Tianyou Chai} (IEEE Life Fellow)
received the Ph.D. degree in control theory and engineering from Northeastern University, Shenyang, China, in 1985. He has been with the Research Center of Automation, Northeastern University, Shenyang, China, since 1985, where he became a Professor in 1988 and a Chair Professor in 2004. His current research interests include adaptive control, intelligent decoupling control, integrated plant control and systems, and the development of control technologies with applications to various industrial processes. Prof. Chai is a member of the Chinese Academy of Engineering, an academician of International Eurasian Academy of Sciences, IEEE Fellow and IFAC Fellow. He is a distinguished visiting fellow of The Royal Academy of Engineering (UK) and an Invitation Fellow of Japan Society for the Promotion of Science (JSPS).
\end{IEEEbiography}

\end{document}